\documentclass[12pt]{article}
\newcommand{\pmb}[1]{{\setbox0=\hbox{#1}%
		\kern-.025em\copy0\kern-\wd0
		\kern.05em\copy0\kern-\wd0
		\kern-.025em\raise.0433em\box0 }}
\newcommand{\fg}[1]{\mbox{\pmb{$#1$}}}

\usepackage{amssymb}
\usepackage{amsmath}
 
\usepackage{multirow}
\usepackage{latexsym}
\usepackage[pdftex]{graphicx}
\DeclareGraphicsExtensions{.pdf,.jpg}
\usepackage{epsf}
\usepackage{graphicx,epsfig}
\usepackage{epstopdf}
\usepackage{amssymb}
\usepackage{amsmath}
\usepackage{subfigure}
\usepackage{setspace}
\usepackage{color}
\usepackage{latexsym}
\usepackage{comment}
\usepackage[pdftex]{graphicx}
\usepackage[colorlinks,citecolor=blue]{hyperref}
\DeclareGraphicsExtensions{.pdf,.jpg}
\usepackage{multirow}
\usepackage{epsf}
\usepackage{array}
\usepackage{booktabs}
\usepackage{authblk}
\usepackage{mathrsfs}
\usepackage{mathrsfs}
\usepackage{accents}
\usepackage{stackengine}
\stackMath
\newcommand\tenq[2][1]{%
	\def\useanchorwidth{T}%
	\ifnum#1>1%
	\stackunder[0pt]{\tenq[\numexpr#1-1\relax]{#2}}{\scriptscriptstyle\sim}%
	\else%
	\stackunder[1pt]{#2}{\scriptscriptstyle\sim}%
	\fi%
}
\newcommand{\myb}[1]{\mbox{\boldmath $#1$}}      
\def\sig{\myb{\sigma}}

\def\om{\myb{\omega}}
\def\tenp{\myb{\otimes}} 
 
\makeatletter
\newcommand*{\rom}[1]{\expandafter\@slowromancap\romannumeral #1@}
\makeatother
\newcommand{\bey}{\begin{eqnarray}}
	\newcommand{\eey}{\end{eqnarray}}

\newcommand{\fvep}{\fg \varepsilon}
\newcommand{\fsg}{\fg \sigma}

\newcommand{\bec}{\begin{center}}
	\newcommand{\eec}{\end{center}}

\usepackage[round]{natbib} 
\begin{document}
	\pagestyle{myheadings}
	\setcounter{tocdepth}{2}
	\baselineskip22pt
	\belowdisplayskip11pt
	\belowdisplayshortskip11pt
	\renewcommand{\thefootnote}{\fnsymbol{hello}}
	\author[1]{Arunabha M. Roy\thanks{Corresponding author | first version submitted on December, 2021}}
	\author[2]{Rikhi Bose}
	\affil[1]{\it Aerospace Engineering, University of Michigan, Ann Arbor, MI  48109, U.S.A.}
	\affil[2]{\it Mechanical Engineering, Johns Hopkins University, Baltimore, MD 21218, U.S.A.}
	\title{\large \bf  Physics-aware deep  learning framework for linear elasticity }
	\date{}
	\maketitle
	
	\bec
	{\bf Abstract}
	\eec
	The paper presents an efficient and robust data-driven deep learning (DL) computational framework developed for linear continuum elasticity problems. 
	The methodology is based on the fundamentals of the Physics Informed Neural Networks (PINNs). 
	For an accurate representation of the field variables, a multi-objective loss function is proposed. 
	It consists of terms corresponding to the residual of the governing partial differential equations (PDE), constitutive relations derived from the governing physics, various boundary conditions, and data-driven physical
	knowledge fitting terms across randomly selected collocation points in the problem domain. 
	To this end, multiple densely connected independent artificial neural networks (ANNs), each approximating a field variable, are trained to obtain accurate solutions. 
	Several benchmark problems including the Airy solution to elasticity and the Kirchhoff-Love plate problem are solved. 
	Performance in terms of accuracy and robustness illustrates the superiority of the current framework showing excellent agreement with analytical solutions. 
	The present work combines the benefits of the classical methods depending on the physical information available in analytical relations with the superior capabilities of the DL techniques in the data-driven construction of lightweight, yet accurate and robust neural networks. 
	The models developed herein can significantly boost computational speed using minimal network parameters with easy adaptability in different computational platforms. 
	\\
	\\
	{\bf 1. Introduction :} 
	\\
	\\
	In recent years, driven by the advancement of bigdata-based  architectures \citep{khan2022sql},  deep  learning (DL) techniques \citep{lecun2015deep} have shown great promises in computer vision \citep{voulodimos2018deep,roy2021deep,roy2022fast,roy2022real,roy2022computer}, object detection \citep{zhao2019object,chandio2022precise,roy2022wildect,singh2023deep}, image classification \citep{rawat2017deep,irfan2021role,jamil2022distinguishing,khan2022introducing}, damage detection \citep{guo2022damage,glowacz2022thermographic,glowacz2021fault} 
	brain-computer interfaces \citep{roy2022efficient,roy2022adaptive,roy2022multi,singh2023understanding} and  
	across various scientific applications
	\citep{butler2018machine,ching2018opportunities,bose2022accurate}.

	The success of these methods, such as various classes of Neural Networks (NNs), can be largely attributed to their capacity in excavating large volumes of data in establishing complex high-dimensional non-linear relations between input features and output \citep{kutz2017deep}. 
	However, the availability of sufficient data is a major bottleneck for analyzing various complex physical systems \citep{butler2018machine,ching2018opportunities}. 
	Consequently, the majority of state-of-the-art machine learning algorithms lack robustness in predicting these systems. 
	Upon availability of sufficient data, these have also garnered considerable success in problems governed by physics, such as dynamical systems \citep{dana2020machine}, geosciences \citep{devries2018deep,bergen2019machine,racca2021robust,saha2021physics,jahanbakht2022sediment}, material science and informatics \citep{butler2018machine,ramprasad2017machine,batra2021emerging,maatta2021gradient}, fluid mechanics \citep{kutz2017deep,brunton2020machine}, various constitutive modeling \citep{tartakovsky2018learning,xu2021learning}, etc. 
	Their applicability however may be further enhanced by utilizing physical information available by mathematical/ analytical means. 
	The recent endeavor of scientific and engineering community has been in attempting to incorporate such physical information within their predictive scheme in small data regimes. 
	\\
	\\
	The incorporation of physical information into the DL framework may have several advantages. 
	First, as previously mentioned, in absence of sufficient data, it may be possible to solely utilize physical knowledge for solving such problems \citep{Raissi-JCP-2019}, or to the least, enhance solutions in a data-driven predictive scheme \citep{Raissi-Science-2020,Karniadakis-Nature-2021}.
	For example, in \cite{sirignano2018dgm}, a  high-dimensional Hamilton–Jacobi–Bellman PDE has been solved by approximating the solution with a DNN  trained to satisfy the differential operator, initial condition, and boundary conditions.
	In incompressible fluid mechanics, the use of the solenoidality condition of the velocity fields restricts the solution space of the momentum equations. 
	Therefore, this condition may be used as a constraint for solving the governing equations (conventional solvers are generally developed in a way to satisfy this constraint through the Poisson equation for pressure), or at least improve the predictions in a data-driven approach. 
	Second, physical systems are often governed by laws that must satisfy certain properties, such as invariance under translation, rotation, reflection, etc. 
	In a purely data-driven approach, it is almost impossible for a DL algorithm to inherit those properties entirely from data without explicit external forcing. 
	Embedding such properties in the DL algorithm might automatically improve the accuracy of the predictions. 
	For example, \citet{ling2016reynolds} used a Tensor-based Neural Network (TBNN) to embed Galilean invariance that improved NN models for Reynolds-averaged Navier Stokes (RANS) simulations for the prediction of turbulent flows. 
	And lastly, any scientific problem is governed by some underlying mechanism dictated by physical laws. 
	Neglect of such physical information in a purely data-driven framework in the current state of affairs is, therefore, an unsophisticated approach, if not an ignorant one.  	
	\\
	\\
	Partial differential equations (PDEs) represent underlying physical processes governed by first principles such as conservation of mass, momentum, and energy. 
	In most cases, analytical solutions to these PDEs are not obtainable. 
	Various numerical methods, such as finite-difference \citep{Sengupta-CUP-2013}, finite element (FE) \citep{Zienkiewicz-FEM-2005}, Chebyshev and Fourier spectral methods \citep{Boyd-spectral-2001}, etc are used to obtain approximate solutions. 
	However, such techniques are often computationally expensive and suffer from various sources of errors due to the complex nature of the underlying PDEs, numerical discretization and integration schemes, iterative convergence techniques, etc. 
	Moreover, the solution of inverse problems is the current endeavor of the engineering community which requires complex formulations and is often prohibitively expensive computationally. 
	The use of the NNs in solving/modeling the PDEs governing physical processes in a forward/ inverse problem is an important challenge worth pursuing, as these methods have the capacity to provide accurate solutions using limited computational resources in a significantly robust framework relative to the conventional methods. 
	In this paper, we explore the possibility of using NN to obtain solutions to such PDEs governing linear continuum elasticity problems applicable in solid mechanics. 
	\\
	\\
	There has been a recent thrust in developing machine learning (ML) approaches to obtain the solution of governing PDEs \citep{Karniadakis-Nature-2021,Von-arXiv-2019}. 
	The idea is to combine traditional scientific computational modeling with a data-driven ML framework to embed scientific knowledge into neural networks (NNs) to improve the performance of learning algorithms \citep{Lagaris-IEEE-1998,Raissi-JCP-2018,Karniadakis-Nature-2021}. 
	The {\it Physics Informed Neural Networks} (PINNs) \citep{Lagaris-IEEE-1998,Raissi-JCP-2019,Raissi-Science-2020} were developed for the solution and discovery of nonlinear PDEs leveraging the capabilities of deep neural networks (DNNs) as universal function approximators achieving considerable success in solving forward and inverse problems in different physical problems such as 
	fluid flows \citep{sun2020surrogate,jin2021nsfnets}, 
	multi-scale flows \citep{lou2021physics},
	heat transfer \citep{cai2021physics,zhu2021machine}, 
	poroelasticity \citep{haghighat2022physics},
	material identification \citep{shukla2021physics}, 
	geophysics \citep{bin2021pinneik,bin2022holistic}, 
	supersonic flows \citep{jagtap2022physics},
	and various other applications \citep{waheed2020eikonal,bekar2022solving}.
	Contrary to traditional DL approaches, PINNs force the underlying PDEs and the boundary conditions in the solution domain ensuring the correct representation of governing physics of the problem. 
	Learning of the governing physics is ensured by the formulation of the loss function that includes the underlying PDEs; therefore labeled data to learn the mapping between inputs and outputs is no more necessary. Such architectural construction  can be utilized  for complex forward and inverse (finding parameters) solutions for various systems of ODEs and 
	PDEs \citep{Karniadakis-Nature-2021}. 
	Additionally, the feed-forward neural networks utilize graph-based automated differentiation (AD) \citep{Baydin-JMLR-2018} to approximate the derivative terms in the PDEs. Various PINNs architectures notably self-adaptive PINNs  \citep{mcclenny2020self}, extended PINNs (XPINN) \citep{hu2021extended, de2022error} have been proposed that demonstrated superior performance. 
	Moreover, multiple DNN-based solvers such as cPINN \citep{jagtap2020conservative},   XPINNs \citep{jagtap2021extended},  and PINNs framework for solid mechanics  \citep{haghighat2021physics} have been developed that provide important advancement in terms of both robustness and 
	faster computation.
	In this regard, \citep{Haghighat-solidmech-2020, haghighat2021physics} have been the breakthrough works geared towards developing a DL-based solver for inversion and surrogate modeling in solid mechanics for the first time utilizing  PINNs theory. 
	Additionally, PINNs have been successfully applied to the solution and discovery in linear elastic solid mechanics \citep{zhang2020physics,samaniego2020energy,Haghighat-2021-nonlocal,Haghighat-2020-energy,Haghighat-2021-Biharmonic, rezaei2022mixed, zhang2022analyses}, elastic-viscoplastic solids \citep{frankel2020prediction,goswami2022physics,arora2022physics,roy2022elastoplastic}, brittle fracture \citep{goswami2020transfer} and computational elastodynamics  \citep{rao2021physics} etc. 
	The solution of PDEs corresponding to elasticity problems can be obtained by minimizing the network's loss function that comprises the residual error of governing PDEs and the initial/boundary conditions. 
	In this regard, PINNs can be utilized as a computational framework for the data-driven solution of PDE-based linear elasticity problems that can significantly boost computational speed with limited network parameters. 
	The potential of the PINNs framework in achieving computational efficiency beyond the capacity of the conventional computational methods for solving complex problems in linear continuum elasticity is the main motivation behind the present work. 
	\\
	\\
	In the present work, an efficient data-driven deep learning computational framework has been presented based on the fundamentals of PINNs for the solution of the linear elasticity problem in continuum solid mechanics. 
	In order to efficiently incorporate physical information for the elasticity problem, an improved multi-objective loss considering additional physics-constrained terms has been carefully formulated that consists of the residual of governing PDE, various boundary conditions, and data-driven physical knowledge fitting terms that demonstrate the efficacy of the model by accurately capturing the elasticity solution. 
	Several benchmark problems including the Airy solution to an elastic plane-stress problem for an end-loaded cantilever beam and simply supported rectangular Kirchhoff-Love thin plate under transverse sinusoidal loading conditions have been solved which illustrates the superiority of the proposed model in terms of accuracy and robustness by revealing excellent agreement with analytical solutions. 
	The employed models consist of independent multi-layer ANNs that are separately trained on minimizing the prescribed loss function specific to the problem under consideration. 
	The performance of PINNs has been evaluated for different activation functions and network architectures. 
	Furthermore, we have illustrated the applicability of data-driven enhancement using the smart initialization of a data-driven learning-based approach in reducing training time, while simultaneously improving the accuracy of the model which is not possible in conventional numerical algorithms. 
	Such an approach would be important in achieving computational efficiency beyond the capacity of conventional computational methods for solving complex linear elasticity problems. 
	The present study also demonstrates the contribution of analytical solutions for the data-driven construction of an accurate and robust PINNs framework that can significantly boost computational speed utilizing minimal trainable network parameters. 
	
	The paper is organized as follows: Section 2 introduces the background of PINNs theory and the generalized idea of implementing multi-objective loss function into the PINNs framework; 
	In section 3, a brief overview  of the theory of linear  elasticity has been   presented;   
	Section 4 introduces the extension of the proposed PINNs framework  for the   Airy  solution to an elastic plane-stress  problem for  an end-loaded cantilever beam; 
	in section 5, the proposed PINNs framework has been extended to the solution of  Kirchhoff–Love thin plate governed by   Biharmonic PDE;  Section 7 deals with the relevant
	finding and prospects of the current work. Finally, the conclusions have been discussed in section 8.
	\\
	\\
	{\bf 2. Physics-Informed Neural Networks :}
	\\
	\\
	The concept of training a NN in the PINNs framework is the construction of the loss function. 
	The loss function is intended to embed the underlying physics which is represented in mathematical terms by the PDEs and the associated boundary conditions. 
	In this section, we discuss the construction of the proposed multi-object loss functions for embedding a data-driven physical model
	that has been associated with the PINNs framework. 
	
	Let us consider a fully connected NN defined by 
	\begin{equation}
		\mathscr{N}^{k+1} (\mathscr{N}^{k})= \varkappa^k (\fg{W}^k \cdot \mathscr{N}^{k}+ \fg{b}^k)
		\label{E-1} 
	\end{equation}
	where $k \in \left\{0, 1, \cdots, N\right\}$ represents  the layer number  of NN.  
	$\mathscr{N}$ is a  nonlinear map defined by 
	$\mathscr{N}^m(\hat{\fg{x}}^m):= \varkappa^m(\fg{W}^m \cdot \fg{x}^m + \fg{b}^m )$ for $m^{th}$-layer where   $\fg{W}^m$ and  $\fg{b}^m$ represents the weights and biases of this transformation, respectively;   $\varkappa (\cdot)$  is the non-linear transformer or activation function acting on a vector element-wise. Therefore, $k=0$ represents the input layer of the NN taking in the input $\fg{x}^0$.  
	

	Also consider a steady state general nonlinear partial differential operator $\mathscr{G}$ operated on a scalar solution   variable $\phi(\vec{x})$ such that, 
	\begin{equation}
		\mathscr{G} \phi(\vec{x})=0 \quad \quad \quad  \vec{x}  \in  \mathbb{R}^{n_{dim}}
		\label{E-2}
	\end{equation}
	Since $\mathscr{G}$ is a differential operator, in general, Eq. \ref{E-2} is accompanied by appropriate boundary conditions to
	ensure the existence and uniqueness of a solution. 
	Let us assume, it is subjected to the boundary condition $\mathscr{B}\, \phi(\partial \vec{\Gamma})=\tau(\partial \vec{x})$ on the boundary $\vec{\Gamma}$ in domain $\Omega \in \mathbb{R}^{n_{dim}}$, ${n_{dim}}$ being the spatial dimension. 
	In a PINNs framework, the solution to Eq. \ref{E-2}, $\phi(\fg{x})$, subjected to the aforementioned boundary condition may be approximated for an input $\fg{x} = \vec{x}$ by constructing a feed-forward NN expressed mathematically as 
	\begin{equation}
		\hat{\phi} = \mathscr{N}^{N} \circledcirc \mathscr{N}^{N-1} \circledcirc \cdots \circledcirc \mathscr{N}^{0} (\fg{x})
		\label{E-3}
	\end{equation}
	\noindent where $\hat{\phi}$ is the approximate solution to Eq. \ref{E-2}; $\circledcirc$ denotes the general compositional construction of the NN; the input to the NN $\mathscr{N}^{0}:=\fg{x}^0=\vec{x} = (x_1, x_2, \cdots x_{n_{dim}}) $ is the spatial coordinate at which the solution is sought. 
	Following Eq. \ref{E-1} and Eq. \ref{E-3}, if $\fg{W}^i$  and $\fg{b}^i$ are all collected in $\theta=\bigcup_{i=0}^N (\fg{W}^i$,  $\fg{b}^i)$, the  output layer $\mathscr{N}^{N}$ contains the approximate solution $\hat{\phi}(\vec{x})$ to the PDE such that  
	\begin{equation}
		\label{E-4}
		\mathscr{N}^{k+1}= \hat{\phi}\,[\fg{x},\theta] = [\hat{\phi}_1, \hat{\phi}_2, ..., \hat{\phi}_m]
	\end{equation}
	The spatial dependence of $\hat{\phi}$  is implicitly contained in the NN parameter $\theta$.
	In the internal/ hidden layers of NN, several variations of nonlinear transformer or the activation function $\varkappa$ may be used, such as, the hyperbolic-tangent function $\tanh(\xi)$, the sigmoid function $\varkappa (\xi)= 1/(1+e^{-\xi})$, the rectified linear unit (ReLU) $\varkappa (\xi) = \mbox{max} (0, \xi)$, etc. 
	The activation in the final layer is generally taken to be linear for regression-type problems considered here. 
	\\
	\\
	{\bf 2.1 Embedding  constraints in NN :}
	\\
	\\
	This section briefly describes the general idea of embedding linear constraints into NN \citep{Lagaris-IEEE-1998,Zaki-PhysRevE}.
	Let us consider $ \mathbb{U}$ and $ \mathbb{A}$, two complete normed vector spaces, where NN function class  $\mathbb{M}  \subset \mathbb{U} $ need  to be constrained.
	A linear constraint on
	$\phi  \in  \mathbb{M}$ can be expressed as:
	\begin{equation}
		\mathscr{P} \phi(\fg{x})=0, \quad \phi  \in  \mathbb{M}
		\label{E-5}
	\end{equation}
	\noindent where, $\mathscr{P} : \mathbb{U}\rightarrow \mathbb{A}$ expresses a linear operator on $\mathbb{U}$. 
	Generally, a such constraint can be realized  for solving PDEs in most of the DL framework by minimizing the following functional  
	\begin{equation}
		\mathscr{J}_A = \Vert \mathscr{P} \phi \Vert_{\mathbb{A}}, \quad \phi  \in  \mathbb{M}
		\label{E-6}
	\end{equation}
	where $\Vert \centerdot \Vert_{\mathbb{A}}$
	denotes  the norm corresponding to space $\mathbb{A}$.
	It is noteworthy to mention that the aforementioned procedure approximately enforces linear constraint in Eq. \ref{E-5}.
	However, the accuracy of the imposed  constraint relies on the relative weighting between the constraint and
	other objectives involved in the training include the satisfaction of the governing PDEs or the integration of data-driven schemes.
	\\
	\\
	{\bf 2.2 Multiple objective loss functions :}
	\\
	\\
	In order to incorporate physical information of the problem, one of the possibilities is to impose Eq. \ref{E-2} as a {\it hard constraint} in $\fg{x} \in \Omega$ while training the NN on the physical data. 
	Mathematically, such a condition is imposed by formulating a constrained optimization problem which can be expressed as \citep{Krishnapriyan-NeuroIPS-2021},
	\begin{equation}
		\begin{array}{rrclcl}
			\displaystyle \min_{\theta} \Delta_\mathcal{L} (\fg{x}, \theta)   \quad \mbox{s.t.}\quad  \mathscr{G} \phi(\vec{x})=0.
		\end{array}
		\label{E-7}
	\end{equation}
	\noindent where $\Delta_L$ represents data-driven physical knowledge fitting term which includes the imposed initial and boundary conditions. 
	$\mathscr{G} \phi(\vec{x}) $ denotes the constraint corresponding to the residual PDE imposing the governing PDE itself. 
	Thus, it is important to carefully impose appropriate constraints for the NN to realize the underlying physics of the problem.

	In the present work, we propose a multi-objective loss function that consists of residuals of governing PDEs, various boundary conditions, and data-driven physical knowledge fitting terms that can be expressed in the following general form:  
	\begin{equation}
		\Delta_\mathcal{L} (\fg{x}, \theta)= \varphi \Vert \mathscr{G} \phi(\fg{x})-\hat{0}\Vert_{\Omega} + \beta_u\Vert \mathscr{B}^{\,\Gamma_u} \phi-g^{\,\Gamma_u}\Vert_{{\Gamma_u}}+ \beta_t\Vert \mathscr{B}^{\,\Gamma_t} \phi-g^{\,\Gamma_t}\Vert_{{\Gamma_t}}+ \alpha \Vert\phi-\hat{\phi}\Vert_{\Omega}+\cdots
		\label{E-8}
	\end{equation}
	\noindent where, $\Delta_\mathcal{L} (\fg{x}, \theta)$ is the total loss function;
	the symbol $\Vert\circledcirc \Vert$ represents  the mean squared error norm, i.e., $\Vert\bigodot \Vert=MSE (\bigodot)$ for regression type problem; 
	$\Vert \mathscr{G} \phi(\fg{x})-\hat{0}\Vert_{\Omega}$ denotes the residual of the governing differential relation in Eq. \ref{E-2} for $\fg{x} \in \Omega$;
	$\Gamma_u$  and $\Gamma_t$ are the Dirichlet and Neumann boundaries subjected to conditions $\mathscr{B}^{\,\Gamma_u} \phi=g^{\,\Gamma_u}$ and $\mathscr{B}^{\,\Gamma_t} \phi=g^{\,\Gamma_t}$, respectively.  
	The values of $g^{\,\Gamma_u}$ and $g^{\,\Gamma_t}$ are specific to the problem under consideration, and therefore, pre-specified as inputs to the problem/ loss function. 
	Note $\varphi$, $\beta_u$, and, $\beta_t$,  are weights associated with each loss term regularizing the emphasis on each term (the higher the relative value, the more emphasis on satisfying the relation). 
	The remaining task is to utilize standard optimization techniques to tune the parameters of the NN minimizing the proposed objective/ loss function $\Delta_\mathcal{L} (\fg{x}, \theta)$ in Eq. \ref{E-8}.

	However, even with a large volume of training data, such an approach may not guarantee that the NN strictly obeys the conservation/governing equations in Eq. \ref{E-2}. 
	Thus, additional loss terms to fit the observation data can be introduced. 
	Hence, in the proposed objective loss function, additional loss terms such as $\Vert\phi-\bar{\phi}\Vert_{\Omega}$ have been included that represent the data-driven physical knowledge fitting term for the state variable $\phi(\vec{x})$. 
	Here, $\bar{\phi}$ is the true (target) value of $\phi$ provided from either the analytical solution (if available), numerical simulation, or experimental observations. 
	$\alpha$ is the weight associated with the data-driven physical knowledge fitting term for $\phi(\vec{x})$. 
	In the NN approximation, various degrees of differentials of the state variable $\phi(\fg{x})$ (i.e., $\phi^{'}(\fg{x})$, $\phi^{''}(\fg{x}), \cdots$ ) can also be included (if known) for stronger coupling in the data-driven approach. 
	The partial differentials of $\phi(\fg{x})$ may be evaluated utilizing the graph-based automatic differentiation \citep{Baydin-JMLR-2018} with multiple hidden layers representing the nonlinear response in PINNs. 
	Following the same steps, the initial conditions can also be incorporated in Eq. \ref{E-8}. 
	The loss from the initial conditions is not included herein due to the quasi-static nature of the elasticity problem. 
	In a more general case, the additional loss term $\Vert\phi_0-\hat{\phi_0}\Vert_{\Omega}^{t=t_0}$ should be added for the loss contribution from the initial condition.

\begin{figure}
		\noindent
		\centering
		\includegraphics[width=0.7\linewidth]{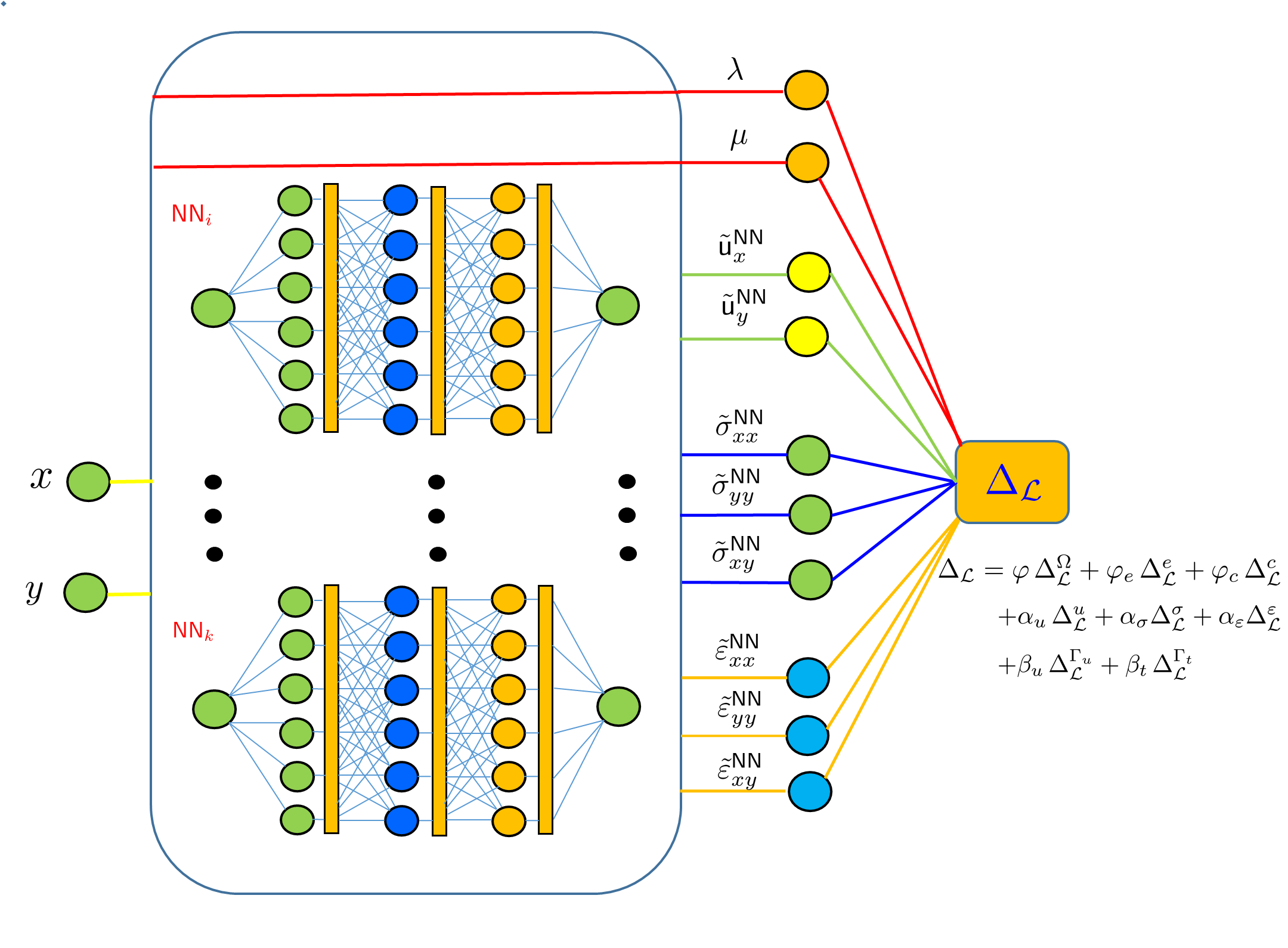}
		\caption{\label{Fig-1} PINNs network architecture for  solving linear elasticity  problem consisting  of multi-ANN ($\mathsf{NN}_i \,\forall\, i=1,k$) for each output variables
			$\tilde{\mathsf{u}}^{\mathsf{NN}}_x(\fg{x})$,
			$\tilde{\mathsf{u}}^{\mathsf{NN}}_y(\fg{x})$,
			$\tilde{\mathsf{\sigma}}^{\mathsf{NN}}_{xx}(\fg{x})$,
			$\tilde{\mathsf{\sigma}}^{\mathsf{NN}}_{yy}(\fg{x})$,
			$\tilde{\mathsf{\sigma}}^{\mathsf{NN}}_{xy}(\fg{x})$, $\tilde{\mathsf{\varepsilon}}^{\mathsf{NN}}_{xx}(\fg{x})$, $\tilde{\mathsf{\varepsilon}}^{\mathsf{NN}}_{yy}(\fg{x})$, and 
			$\tilde{\mathsf{\varepsilon}}^{\mathsf{NN}}_{xy}(\fg{x})$,
			with independent variable  $\fg{x}=(x, y) $ as input features.}
	\end{figure}

	Finally, the optimal network parameters of NN $\tilde{\theta}$ can be obtained by optimizing the loss function in Eq. \ref{E-8} as
	\begin{equation}
		\begin{array}{rrclcl}
			\displaystyle \tilde{\theta}= \arg \min_{\theta  \subset \mathbb{R}^{N^t}} \Delta_\mathcal{L} (\fg{\bar{X}}, \theta).
		\end{array}
	\end{equation}
	\noindent where, $\tilde{\theta}:=\bigcup_{i=0}^N (\fg{\tilde{W}}^i$,  $\fg{\tilde{b}}^i)$ is the set of optimized network parameters; $N^t$ is the total number of trainable parameters; and $\fg{\bar{X}} \in \mathbb{R}^{N_c \times N^t} $ is the set of $N_c$ collocation points used for optimization. 
	\\
	\\
	{\bf 3 Theory of linear elastic solid:}
	\\
	\\
	Consider an undeformed configuration $\mathcal{B}$ of an elastic body bounded in the domain $\Omega  \subset \mathbb{R}^{n_{dim}}$ $(1 \leq {n_{dim}} \leq 3)$ with boundary $\Gamma=\Gamma_u \cup \Gamma_t$ where $\Gamma_u  \ne \emptyset$   is the Dirichlet boundary, $\Gamma_t$ is the Neumann boundary, and $\Gamma_u \cap \Gamma_t= \emptyset$. 
	With respect to the undeformed surface, the elastic body can be subjected to a  prescribed displacement $\bar{{\fg u}}$ on $\Gamma_D$, and a prescribed surface traction $\bar{{\fg t}} \in [\mathcal{L}^2(\Gamma_t )]^{n_{dim}}$. 
	Additionally, a body force  of density ${\fg B}  \in  [\mathcal{L}^2(\Omega )]^{n_{dim}}$ in $\Omega$ ­can be  prescribed with respect to the undeformed volume.
	Using a standard basis $  {\left\{ \fg{e_i} \right\}}$ in $\mathbb{R}^{n_{dim}}$, we can express the displacement, $\fg {u}=u_i \fg {{e_i}}$, and its gradient, $\nabla \fg {u} = \frac{1}{2} \left(u_{i,j}+u_{j, i}\right) \fg {e_i}\, \tenp\, \fg {e_j}$; where, $\tenp$ denotes the tensor products. 
	Second order symmetric tensors are linear transformations in $\mathbb{S}$, defined as $\mathbb{S}:= \left\{ \fg {\xi} : \mathbb{R}^{n_{dim}} \rightarrow \mathbb{R}^{n_{dim}} |\, \fg {\xi}= \fg {\xi^T} \right\}$ with inner product $\fg {\xi}\, :\, \fg {\xi}=$ tr $\left[ \fg {\xi}\fg {\xi^T}\right]  \equiv \xi_{ij} \xi_{ij} $. 
	Therefore, the stress tensor can be expressed as ${\fsg} := \sigma_{ij}\fg {e_i}\, \tenp\, \fg {e_j}$.
	For infinitesimal strain,  displacement gradient tensor $\nabla \fg{u}$ can be expressed as:  $\nabla \fg{u}=\fvep+\mathbf{\om}$ where \,  $\fvep: = \frac{1}{2}\left[\nabla \fg{u}+ \nabla (\fg{u})^{\mathsf{T}}\right]$ is the infinitesimal strain tensor  with $\nabla\times \fvep =e_{ijk}\, \varepsilon_{rj, i}\,\fg{e_k}\, \tenp\, \fg{e_r}$, and  $\mathbf{\om}: = \frac{1}{2}\left[\nabla \fg{u}- \nabla (\fg{u})^{\mathsf{T}}\right]$  is the  infinitesimal rotation tensor.
	\\
	\\
	{\bf 3.1  Compatibility condition:}
	\\
	\\
	In the context of infinitesimal strain theory, we seek to find 
	${\fg u} :\Omega \rightarrow \mathbb{R}^{n_{dim}}$ and corresponding  $\fvep :\Omega \rightarrow \mathbb{R}^{{n_{dim}} \times {n_{dim}}}$, and  $\fsg :\Omega \rightarrow \mathbb{R}^{{n_{dim}} \times {n_{dim}}}$ for a given infinite elastic solid satisfying the following compatibility conditions  \citep{marsden1994mathematical}: 
	\begin{equation}
		\fg{R :}=\nabla \times (\nabla\times \fvep)^{\mathsf{T}}=\fg{0};
		\label{E-10} 
	\end{equation}
	\noindent where, $\fg{R}$ is Saint-Venant compatibility tensor.
	Alternatively, the elastic solid should satisfy  the Navier–Cauchy equations which can be expressed as \citep{lurie2010theory}:    
	\begin{equation}
		\begin{aligned}
			(\lambda +\mu)\mathbf{\nabla }&  (\mathbf{\nabla \cdot {\fg u}}) +\mu \mathbf{\Delta {\fg u}}+\mathbf {\fg B}={\fg 0}, \quad \,\, \mbox{in}\,\, \Omega \\
			& \fg {u} \mid_{{\Gamma_D}} \,= \fg {\bar{u}}; 
		\end{aligned}
		\label{E-11}
	\end{equation}
	where $\fg {u}= (u_1, u_2, ..., u_{n_{dim}})$ is the unknown displacement field; 
	$\mu >0$ and $\lambda> -\mu$ are Lame constants; $\mathbf{\nabla }$, $\mathbf{\Delta}$, and $\mathbf{\nabla}$ represent the gradient, the Laplacian, and the divergence operators, respectively. Equation \ref{E-11} satisfies the continuity of the displacement field $\fg{u}$ and  Dirichlet boundary condition. 
	\\
	\\
	{\bf 3.2  Equilibrium condition:}
	\\
	\\ 
	In addition,  the equilibrium condition  and the Neumann boundary condition should be satisfied  which can be expressed as \citep{marsden1994mathematical}:
	\begin{equation}
		\begin{aligned}
			\mathbf{\nabla \cdot} {\,\fsg} & +\mathbf {\fg B}={\fg 0}, \quad \,\, \mbox{in}\,\, \Omega \\
			& {\fg t}:=\mathbb{T} \fg{ u}=\bar{{\fg t}}, \quad \,\, \mbox{on}\,\, \Gamma_t  \quad \quad 
			{\fsg}\mid_{{\Gamma_t}} \, \fg {\hat{n}} = \fg {\bar{t}}
			\label{E-12}
		\end{aligned}
	\end{equation}
	\noindent where, $\bar{{\fg t}}$ is a prescribed function on $\Gamma_t$; $\fg {\hat{n}}$ is the field normal to ${\Gamma_t}$. 
	Equation \ref{E-12} satisfies the momentum equation and the Neumann boundary condition where $\mathbb{T}$ follows the conformal derivative operator such that \citep{atkin2005introduction} 
	\begin{equation}
		\mathbb{T} \fg {u} = \lambda (\mathbf{\Delta}\, \fg {u}) \cdot \fg {\hat{n}} +2 \mu \frac{\partial \fg{u} }{\partial \fg {\hat{n}}}+ \mu \, \fg {\hat{n}} \times (\nabla\times \fg {u})
		\label{E-13}
	\end{equation}
	\\
	{\bf 3.3  Constitutive relation:}
	\\
	\\
	Subsequently, the  elastic constitutive relation can be expressed  from generalized Hooke's law \citep{timoshenko1970theory} as:
	\begin{equation}
		\fsg=  \fg {C} : \fvep
		\label{E-14}
	\end{equation}
	\noindent where, the fourth-order stiffness tensor $\fg{C} =C_{ijkl}\, \fg {e_i}\, \tenp\, \fg {e_j}\,\tenp \, \fg {e_k}\, \tenp \, \fg {e_l}$ denotes the constitutive relation that maps the displacement gradient $\mathbf{\nabla} \fg {u}$ to the Cauchy stress tensor $\fsg$. 
	For an isotropic linearly elastic material, $C_{ijkl}= \lambda\delta_{ij}\delta_{kl}+ \mu (\delta_{ik}\delta_{jl}+\delta_{il}\delta_{jk}) $ where  $\delta_{ij}$ is the Kronecker delta. 
	The components of the stress tensor $\fsg$, and the strain tensor $\fvep$, are expressed as : 
	\begin{equation}
		\sigma_{ij}(\fg{u})= \lambda\delta_{ij} \sum_{k=1}^{{n_{dim}}}\varepsilon_{kk}(\fg{u})+2 \mu \varepsilon_{ij}(\fg{u}), \quad \quad \varepsilon_{ij} (\fg{u})=\frac{1}{2} \left(\frac{\partial u_i}{\partial x_j} +\frac{\partial u_j}{\partial x_i}\right), \quad \quad i,j=1, 2, ..., {n_{dim}}.
		\label{E-15}
	\end{equation}
	Note that $\sig$ is the Cauchy stress tensor in linear elasticity applicable under small deformation. 
	The constitutive relation in terms of strain can be alternatively expressed as,
	\begin{equation}
		\varepsilon_{ij,kl}+\varepsilon_{kl,ij}-\varepsilon_{ik,jl}-\varepsilon_{jl,ik}=0 \quad \quad i, j, k, l \in 1, 2, ..., {n_{dim}}.
		\label{E-16}
	\end{equation}
	Equations governing a linear elastic boundary value problem (BVP) are defined by Eqs. \ref{E-11}--\ref{E-16} where the field variables  $\fg{u}, \sig, \fvep$ can be obtained for given material constants \citep{atkin2005introduction,lurie2010theory}.
	\\
	\\
	{\bf 4. PINNs formulation for continuum linear elasticity:}
	\\
	\\
	The proposed PINNs framework is applied to linearly elastic solids. 
	A two-dimensional ($n_{dim}=2$) problem is considered. 
	The input features (variables) to the models are the spatial coordinates $\fg{x}= (x, y)$. 
	A separate NN is used to approximate each output field variable. 
	As shown in Fig. \ref{Fig-1}, displacement $\fg{u}(\fg{x})$, stress $\fsg (\fg{x})$, and strain $\fvep (\fg{x})$ fields are obtained by densely connected independent ANNs. 
	For $n_{dim}=2$, considering symmetry of the stress and strain tensors, $\fg{u}(\fg{x})$, $\fsg (\fg{x})$, and $\fvep (\fg{x})$ fields can be approximated as:  
	\begin{equation}
		\fg{u}(\fg{x})  \simeq \Xi^{\mathsf{NN}}_{\fg{u}}({\fg{x}}) =
		\begin{bmatrix}
			\tilde{\mathsf{u}}^{\mathsf{NN}}_x(\fg{x})  \\
			\tilde{\mathsf{u}}^{\mathsf{NN}}_y(\fg{x})
		\end{bmatrix} 
		\label{E-17}
	\end{equation}
	\begin{equation}
		\fsg (\fg{x})  \simeq \Xi^{\mathsf{NN}}_{\fsg} (\fg{x}) =
		\begin{bmatrix}
			\tilde{\mathsf{\sigma}}^{\mathsf{NN}}_{xx}(\fg{x}) & \tilde{\mathsf{\sigma}}^{\mathsf{NN}}_{xy}(\fg{x}) \\
			\tilde{\mathsf{\sigma}}^{\mathsf{NN}}_{yx}(\fg{x}) & \tilde{\mathsf{\sigma}}^{\mathsf{NN}}_{xy}(\fg{x})
		\end{bmatrix};\quad\quad \fvep (\fg{x})  \simeq \Xi^{\mathsf{NN}}_{\fvep} (\fg{x}) = 
		\begin{bmatrix}
			\tilde{\mathsf{\varepsilon}}^{\mathsf{NN}}_{xx}(\fg{x}) & \tilde{\mathsf{\varepsilon}}^{\mathsf{NN}}_{xy}(\fg{x}) \\
			\tilde{\mathsf{\varepsilon}}^{\mathsf{NN}}_{yx}(\fg{x}) & \tilde{\mathsf{\varepsilon}}^{\mathsf{NN}}_{xy}(\fg{x})
		\end{bmatrix}
		\label{E-18}
	\end{equation}
	Here $\Xi^{\mathsf{NN}}_{\fg{u}}({\fg{x}})$, $\Xi^{\mathsf{NN}}_{\fsg} (\fg{x})$, and $\Xi^{\mathsf{NN}}_{\fvep} (\fg{x})$ denote the NN approximations for $\fg{u}(\fg{x})$, $\fsg (\fg{x})$, and  $\fvep (\fg{x})$, respectively. 
	\\
	\\
	{\bf 4.1 Loss function:}
	\\
	\\
	To define the loss function for the linear elasticity problem, governing equations including compatibility conditions, equilibrium conditions, constitutive relations, and boundary conditions that fully describe the problem have been considered. 
	Additionally, as in a data-driven approach, the field variables in Eq. \ref{E-8} have been included. 
	The generalized mutli-objective loss functional $\Delta_\mathcal{L}$ can be expressed as: 
	\begin{equation}
		\Delta_\mathcal{L} (\fg{x}, \theta)=\varphi\, \Delta^{{\Omega}}_\mathcal{L}+
		\varphi_e\,  \Delta^{e}_\mathcal{L}+
		\varphi_c\,  \Delta^{c}_\mathcal{L}+ 
		\beta_u\, \Delta^{{\Gamma_u}}_\mathcal{L}+\beta_t\, \Delta^{{\Gamma_t}}_\mathcal{L}+\alpha_{{\fg u}}\, \Delta^{{\fg u}}_\mathcal{L}+\alpha_{{\fsg}} \Delta^{{\fsg}}_\mathcal{L}+\alpha_{{\fvep}}\Delta^{{\fvep}}_\mathcal{L}
		\label{E-19}
	\end{equation}
	\noindent where, $\Delta^{e}_\mathcal{L}$, $\Delta^{c}_\mathcal{L}$, and $\Delta^{{\Omega}}_\mathcal{L}$ are the loss components from the equilibrium condition (Eq. \ref{E-12}), constitutive relation (Eq. \ref{E-14}), and the compatibility condition (Eq. \ref{E-15}), respectively; $\Delta^{{\Gamma_u}}_\mathcal{L}$ and  $\Delta^{{\Gamma_t}}_\mathcal{L}$ represent the  loss components computed at the Dirichlet boundary $\Gamma_u$, and the Neumann boundary $\Gamma_t$ (Eq. \ref{E-11}), respectively; $ \Delta^{{\fg u}}_\mathcal{L}$, $\Delta^{{\fsg}}_\mathcal{L}$, and $\Delta^{{\fvep}}_\mathcal{L}$ are the loss components for the fields $\fg{u}(\fg{x})$, $\fsg (\fg{x})$, and $\fvep (\fg{x})$, respectively, when a data driven approach is pursued. 
	The coefficients $\varphi, \varphi_e, \varphi_c, \beta_u, \beta_t, \alpha_{{\fg u}}, \alpha_{{\fsg}}$, and $\alpha_{{\fvep}}$ are the weights associated with each loss term that dictates the emphasis on each penalty term. 
	Evidently, the terms in the cost function are the measures of the errors in the displacement and stress fields, the momentum balance, and the constitutive law. 
	The explicit expression for each term in $\Delta_\mathcal{L} (\fg{x}, \theta)$ is, 
	\bey
	\Delta^{{\Omega}}_\mathcal{L} &=& \frac{1}{N_c^{\Omega}} \sum_{l=1}^{N_c^{\Omega}} \lVert \mathbf{\nabla \cdot\Xi^{\mathsf{NN}}_{\sig} (\fg{x}}_{l \vert \Omega})  +\fg{B}(\fg{x}_{l \vert \Omega}) \rVert 
	\label{E-20}\\
	\Delta^{c}_\mathcal{L} &=& \frac{1}{N_c^{\Omega}} \sum_{l=1}^{N_c^{\Omega}} \lVert \mathbf{\Xi^{\mathsf{NN}}_{\fsg} (\fg{x}}_{l \vert \Omega})  -\fg{C} \left[\nabla \cdot \Xi^{\mathsf{NN}}_{\fg{u}}(\fg{x}_{l \vert \Omega})\right] \rVert
	\label{E-21}\\
	\Delta^{{\Gamma_u}}_\mathcal{L}&=& \frac{1}{N_c^{\Gamma_u}} \sum_{k=1}^{N_c^{\Gamma_u}} \lVert \mathbf{\Xi^{\mathsf{NN}}_{\fg{u}} (\fg{x}}_{k \vert \Gamma_u})  -\fg{\bar{u}}(\fg{x}_{k \vert \Gamma_u})| \rVert
	\label{E-22}\\
	\Delta^{{\Gamma_t}}_\mathcal{L}&=& \frac{1}{N_c^{\Gamma_t}} \sum_{j=1}^{N_c^{\Gamma_t}} \lVert \mathbf{\Xi^{\mathsf{NN}}_{\fsg}} (\fg{x}_{j \vert \Gamma_t}){\fg {\hat{n}}}  -\fg {\bar{t}}(\fg{x}_{j \vert \Gamma_t}) \rVert
	\label{E-23}\\
	\Delta^{{\fg u}}_\mathcal{L}&=& \frac{1}{N_c^{\Omega}} \sum_{l=1}^{N_c^{\Omega}} \lVert \mathbf{ \Xi^{\mathsf{NN}}_{\fg{u}} (\fg{x}}_{l \vert \Omega})  -\hat{\fg{u}}(\fg{x}_{l \vert \Omega}) \rVert 
	\label{E-24}\\
	\Delta^{{\fsg}}_\mathcal{L}&=&\frac{1}{N_c^{\Omega}} \sum_{l=1}^{N_c^{\Omega}} \lVert \mathbf{ \Xi^{\mathsf{NN}}_{\fsg} (\fg{x}}_{l \vert \Omega})  -\hat{\fsg} (\fg{x}_{l \vert \Omega}) \rVert  
	\label{E-25}\\
	\Delta^{{\fvep}}_\mathcal{L}&=& \frac{1}{N_c^{\Omega}} \sum_{l=1}^{N_c^{\Omega}} \lVert \mathbf{ \Xi^{\mathsf{NN}}_{\fvep} (\fg{x}}_{l \vert \Omega})  -{\hat{\fvep}}(\fg{x}_{l \vert \Omega}) \rVert 
	\label{E-26}
	\eey
	\noindent where, $\left\{\fg{x}_{1 \vert \Omega}, ..., \fg{x}_{N_c^{\Omega} \vert \Omega}\right\}$ are randomly chosen collocation points over the domain $\Omega$; $\left\{\fg{x}_{1 \vert {\Gamma_u}}, ..., \fg{x}_{N_c^{{\Gamma_u}} \vert {\Gamma_u}}\right\}$ and $\left\{\fg{x}_{1 \vert {\Gamma_t}}, ..., \fg{x}_{N_c^{{\Gamma_t}} \vert {\Gamma_t}}\right\}$ are those chosen randomly along the boundaries $\Gamma_u$ and $\Gamma_t$, respectively. 
	The terms $\hat{\fg{u}}(\fg{x}_{l \vert \Omega})$, $\hat{\fsg} (\fg{x}_{l \vert \Omega})$, and ${\hat{\fvep}}(\fg{x}_{l \vert \Omega})$ represent the true (target) value obtained by means of analytical solution or high-fidelity simulation. 
	The weights $\varphi, \varphi_e, \varphi_c \in  \mathbb{R}^+$ are  the weights corresponding to the compatibility, equilibrium,  and constitutive relations, respectively. 
	In general, these coefficients can be prescribed as 1  for solving a relatively less complex problem, whereas, $ \beta_u$ and $\beta_t$ are the binary (i.e., either 0 or 1) integers. 
	The weights $\alpha_i=1 ; \,\,\forall\,\, i={\fg u}, {\fsg}, \fvep$  for a complete  data driven approach for $\fg{u}(\fg{x})$, $\fsg (\fg{x})$, and $\fvep (\fg{x})$, respectively at the collocation points $N_c^{\Omega}$. 
	However, we prescribe $\alpha_i =0\,\, \forall\,\,  (i={\fg u}, {\fsg}, \fvep)$ as labeled training data is unavailable, which may not guarantee the accuracy of PINNs solutions. 
	
\begin{figure}
	\noindent
	\centering
	\includegraphics[width=1.1\linewidth]{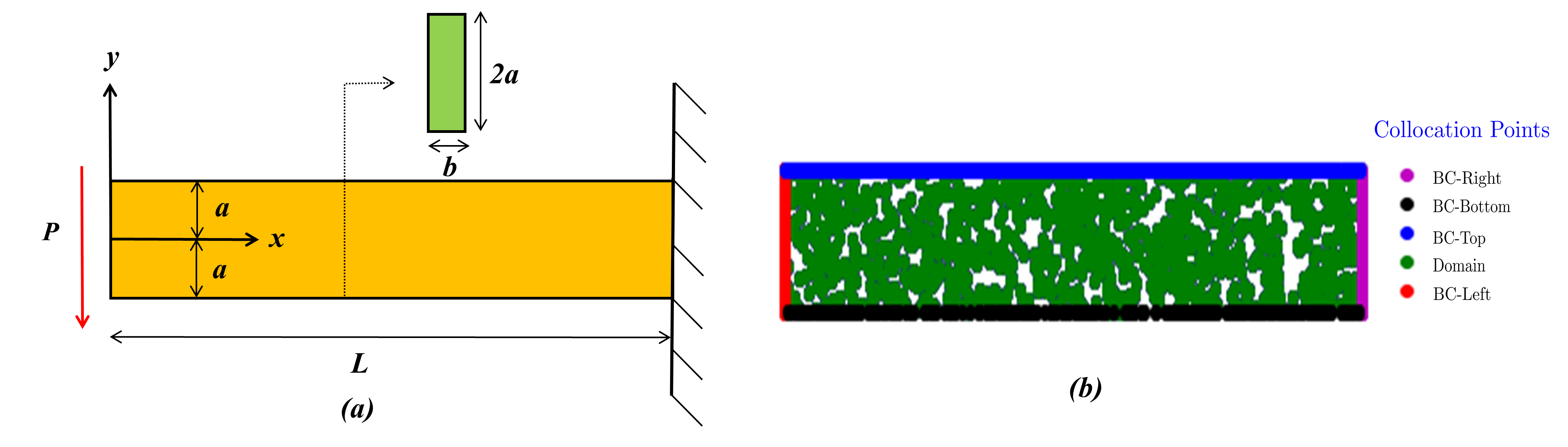}
	\caption{\label{Fig-2} (a) Elastic plane-stress problem for an end-loaded cantilever beam of length $L$, height $2a$ and out-of-plane thickness $b$ which has been clamped at $x=L$; (b) distributions of total collocations points $N_c=5,000$  on the problem domain and various boundaries during PINNs training. }
\end{figure}

	The forward problem is studied herein, where the displacement, stress, and strain fields are obtained as the PINNs solutions assuming material properties  $\lambda$ and $\mu$ remain constant. 
	However, the loss functional in Eq. \ref{E-19} can also be utilized in an inverse problem for parameter identification, where $\lambda$ and $\mu$ can be treated as network outputs which may vary during training (Fig. \ref{Fig-1}). 
	For the network construction in the PINNs framework, SciANN \citep{Haghighat-CMAME-2021}, a convenient high-level Keras \citep{Chollet-keras-2015} wrapper for PINNs is used. 
	\\
	\\
{\bf 4.2 Solution for linear elasticity problem :}
\\
\\
For this study, an end-loaded isotropic linearly elastic cantilever beam of height $2a$, length $L$, thickness $b$ (assuming $b\ll a$)  has been considered to ensure a state of plane-stress condition as shown in Fig. \ref{Fig-2}. 
The left edge of the beam is subjected to a resultant force $P$. 
Whereas, the right-hand end is clamped.
The top and bottom surfaces of the beam, $y=\pm a$ are traction free. 
An approximate solution to the problem can be obtained from the Airy function discussed next. 
\\
\\
{\bf 4.2.1 The Airy solution to the end-loaded cantilever beam:}
\\
\\
The Airy solution in Cartesian coordinates $\Omega  \subset \mathbb{R}^2$ can be found from the Airy potential $\phi(x ,y)$ that satisfies \citep{bower2009applied}, 
\begin{equation}
	\mathbf{\nabla} \phi= \frac{\partial^4 \phi}{\partial x^4} +2\frac{\partial^4 \phi}{\partial x^2 \partial y^2}+\frac{\partial^4 \phi}{\partial y^4}= \mathbf{C}(\nu)(\frac{\partial b_x}{\partial x}+\frac{\partial b_y}{\partial y})
	\label{E-27}
\end{equation}
\noindent where,  
\begin{equation}
	\mathbf{C}(\nu) = \left\{
	\begin{array}{ll}
		\frac{1-\nu}{1-2\nu} & \mbox{(plane strain)} \\
		\frac{1}{1-\nu} & \mbox{(plane stress)}
	\end{array}
	\right.
	\label{E-28}
\end{equation}
Here, the body forces $b_x$, $b_y$ have the form  $\rho_0b_x=\frac{\partial \Omega}{\partial x}, \, \rho_0b_y=\frac{\partial \Omega}{\partial y}$ ;  $\Omega(x , y)$ is the positional  scalar function. 
The solution of the Airy  function can be expressed in the polynomial form  $\phi (x, y)=\sum_{m=0}^{\infty }\sum_{n=0}^{\infty}A_{mn}\, x^m y^n$. 
For $m + n \leq 3$, the terms automatically satisfy the biharmonic equation for any $A_{mn}$. 
Additionally,  $\phi$ must satisfy the following traction boundary conditions on $\Omega$. 
\begin{equation}
	\frac{\partial^2 \phi}{\partial y^2}n_x -\frac{\partial^2 \phi}{\partial x \partial y}n_y=t_x; \quad \frac{\partial^2 \phi}{\partial x^2}n_y -\frac{\partial^2 \phi}{\partial x \partial y}n_y=t_y
	\label{E-29}
\end{equation}
Here, $(n_x,n_y)$ are the components of a unit vector normal to the boundary.
For the end-loaded cantilever beam, the Airy function can be formulated as, 
\begin{equation}
	\phi= -\frac{3P}{4ab} x y+\frac{P}{4a^3b} x y^3
	\label{E-30}
\end{equation}
\noindent where, $\sigma_{xx}= \frac{\partial^2 \phi}{\partial y^2} -\Omega;\quad\sigma_{yy}= \frac{\partial^2 \phi}{\partial x^2} -\Omega;\quad \sigma_{xy}=\sigma_{yx}= -\frac{\partial^2 \phi}{\partial x\partial y}$ with $\Omega=0$. 
At the clamped end, $x_1=L$, displacement boundary conditions are $u_x=u_y=\partial u_y/\partial x=0$. 
The top and bottom surfaces of the beam (i.e., $y=\pm a$) are traction free, $\sigma_{ij}n_i=0$, that requires $\sigma_{yy}=\sigma_{xy}=0$.
Whereas, the resultant of the traction acting on the surface at $x=0$ is $-Pe_y$ with traction vector $t_i=\sigma_{ij}n_{j}=-\sigma_{xy}\delta_{iy}=-\frac{3P}{4ab}(1-\frac{y^2}{a^2})\delta_{iy}$. 
The resultant force can be obtained as :
$F_i=b\int_{-a}^{a}-\frac{3P}{4ab}(1-\frac{y^2}{a^2})\delta_{iy}dx_2=-P\delta_{iy}$. 
On satisfaction of the aforementioned conditions, approximate analytical solutions for the displacements $u_{x}$, $u_{y}$, the strain fields $\varepsilon_{xx}$, $\varepsilon_{yy}$, $\varepsilon_{xy}$ and the stress fields  $\sigma_{xx}$, $\sigma_{yy}$, $\sigma_{xy}$ can be expressed as:
\bey
u_{x} &=& \frac{3P}{4Ea^3b} x^2y-(2+\mu)\frac{P}{4Ea^3b}y^3+3(1+\mu)\frac{Pa^2}{2Ea^3b}y-\frac{3PL^2}{4Ea^3b}y
\label{E-31}\\
u_{y} &=& - \frac{3\mu P}{4Ea^3b}xy^2-\frac{P}{4Ea^3b}x^3+\frac{3PL^2}{4Ea^3b}x-\frac{PL^3}{2Ea^3b}
\label{E-32}\\
\varepsilon_{xx}&=&\frac{3P}{2Ea^3b} x y; \quad \varepsilon_{yy}=-\frac{3P\mu}{2Ea^3b} x y; \quad \varepsilon_{xy}=\frac{3P(1+\mu)}{4Eab}\left(1-\frac{y^2}{a^2}\right)
\label{E-33}\\
\sigma_{xx}&=&\frac{3P}{2a^3b} x y; \quad \sigma_{yy}=0; \quad \sigma_{xy}=\frac{3P}{4ab}\left(1-\frac{y^2}{a^2}\right)
\label{E-34}
\eey
These analytical solutions for $\fg{u}(\fg{x})$, $\fsg (\fg{x})$, and  $\fvep (\fg{x})$ have been used as $\hat{\fg{u}}(\fg{x}_{l \vert \Omega})$, $\hat{\fsg} (\fg{x}_{l \vert \Omega})$, and ${\hat{\fvep}}(\fg{x}_{l \vert \Omega})$ at the collocation points for data-driven enhancement in Eqs. \ref{E-24}-\ref{E-26}, respectively, for solving the field variables in the proposed PINNs framework.
\\
\\	
{\bf 4.2.2  PINNs solutions for linear elasticity problem:}	
\\
\\
For the benchmark, end-loaded cantilever beam problem, $L=3$ m, $a=0.5$ m, and $b =0.001$ m have been considered. 
The material properties are, Young’s modulus $E=1$ GPa, and the Poisson ratio $\nu=0.25$ as shown in Fig. \ref{Fig-2} -(a). 
Unless otherwise stated, a total of $N_c= 5,000$ randomly distributed collocation points over the domain and boundaries have been used for training the PINNs model as shown in Fig. \ref{Fig-2} -(a). 
During training, the optimization loop was run for 500 epochs using the Adam optimization scheme with a learning rate of 0.001, and a batch size of 32 for optimal accuracy and faster convergence.
\begin{figure}
	\noindent
	\centering
	\includegraphics[width=1.1\linewidth]{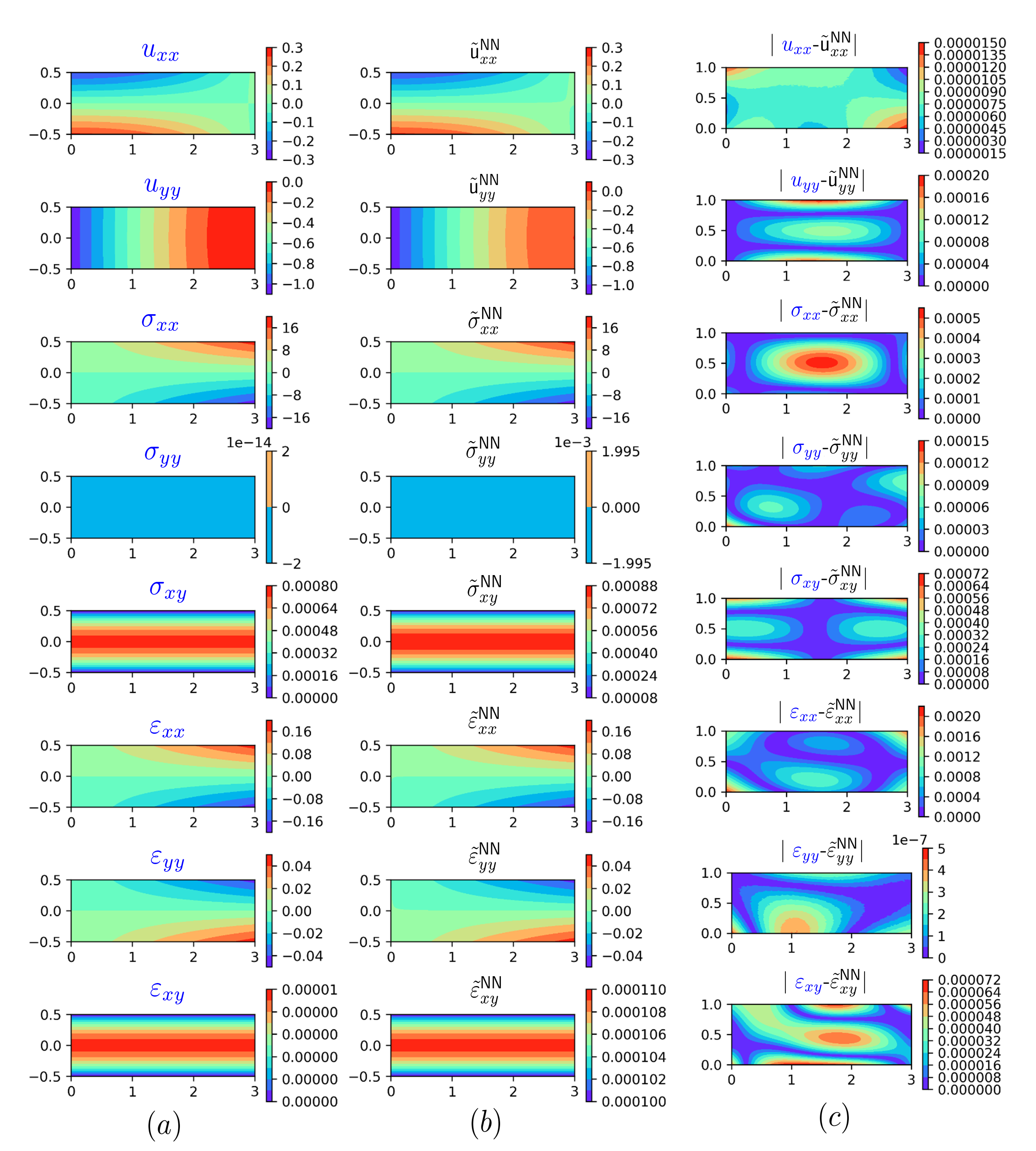}
	\caption{\label{Fig-3} (a) The Airy  solutions for displacements $u_x$, $u_y$, stresses $\sigma_{xx}$, $\sigma_{yy}$, $\sigma_{xy}$, strains $\varepsilon_{xx}$, $\varepsilon_{yy}$, $\varepsilon_{xy}$;
		(b) corresponding PINNs solutions for  	$\tilde{\mathsf{u}}^{\mathsf{NN}}_x$,
		$\tilde{\mathsf{u}}^{\mathsf{NN}}_y$,
		$\tilde{\mathsf{\sigma}}^{\mathsf{NN}}_{xx}$,
		$\tilde{\mathsf{\sigma}}^{\mathsf{NN}}_{yy}$,
		$\tilde{\mathsf{\sigma}}^{\mathsf{NN}}_{xy}$, $\tilde{\mathsf{\varepsilon}}^{\mathsf{NN}}_{xx}$, $\tilde{\mathsf{\varepsilon}}^{\mathsf{NN}}_{yy}$, and 
		$\tilde{\mathsf{\varepsilon}}^{\mathsf{NN}}_{xy}$; (c)  absolute error between the Airy solutions and PINNs predictions associated with each field variables for an end-loaded cantilever beam. } 
\end{figure}

The Airy  solutions   for  various fields including  displacements $u_x$, $u_y$,  stresses $\sigma_{xx}$, $\sigma_{yy}$, $\sigma_{xy}$, and strains $\varepsilon_{xx}$, $\varepsilon_{yy}$, $\varepsilon_{xy}$ as in Eqs. \ref{E-31}-\ref{E-34} are shown in Fig. \ref{Fig-3}-(a). 
The corresponding PINNs approximations using the tanh activation function are shown in  Fig. \ref{Fig-3} -(b). 
Additionally, in Fig. \ref{Fig-3} -(c), the absolute error between the Airy solutions and PINNs predictions for each field variable is shown. 
The overall results from PINNs are in excellent agreement with the Airy solutions. 
The PINNs approximations attained satisfactory accuracy with low absolute errors for all field variables. 
For the displacement fields, the absolute error is relatively high near to clamped edge for $u_x$. 
For $u_y$, the absolute error is maximum at the midsection and near the horizontal edges as shown in Fig. \ref{Fig-3} -(c). 
This is due to the approximate nature of the Airy solutions at clamped end $x_1=L$ for the displacement boundary conditions $u_x=u_y=\partial u_y/\partial x=0$. 
Such differences also propagate through the solutions of stress and strain fields, where PINNs predictions slightly deviate from the Airy solutions, in particular,  near the free vertical and horizontal edges as shown in Fig. \ref{Fig-3} -(c). 
However, according to Saint-Venant’s principle, these deviations do not sufficiently influence the solution far from the end, which is reflected in the result. 
Overall, the proposed PINNs model can capture the distributions of various fields accurately from the solution of the Airy stress function.  
\begin{figure}
	\noindent
	\centering
	\includegraphics[width=0.95\linewidth]{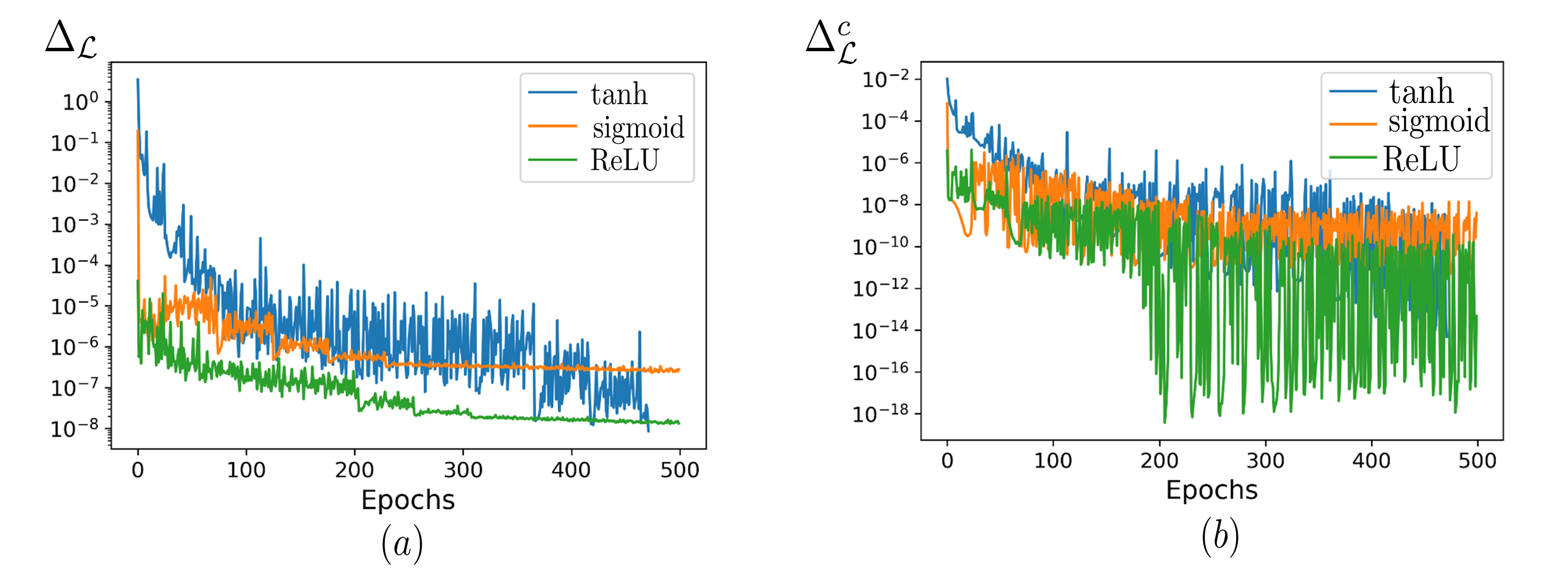}
	\caption{\label{Fig-4} Comparison of (a) total loss $\Delta^{{\Omega}}_\mathcal{L}$; (b) constitutive loss  $\Delta^{{\Omega}}_\mathcal{L}$ for  tanh, sigmoid and ReLU activation functions for  network parameters  $\mathscr{N}=20, L_n=5$.}
\end{figure} 
\\
\\
{\bf 4.2.3 Suitable activation function :}
\\
\\	
The impact of the use of various activation functions on training the PINNs models in predicting field variables and the epoch evolution of various components of the loss function is explored. 
The ReLU, sigmoid, and tanh activation functions are compared; the network architecture remains the same: the number of neurons in each layer  $\mathscr{N}=20$ with the total number of hidden layers  $L_n=5$ in the PINNs model. 
The evolution of the total loss $\Delta_\mathcal{L}$, and the constitutive loss $\Delta^{{\Omega}}_\mathcal{L}$ are depicted in  Fig. \ref{Fig-4}. 
Additionally, values of the various loss components and training times $t_{tr}$ at the end of training are compared in Table. \ref{T-1}. 
Evidently, the tanh activation provides the best performance in terms of the value of the total loss at the end of training. 
The final constitutive loss with tanh activation is significantly lower compared to the other two activations illustrating the suitability of the use of the tanh activation for the PINNs model for solving the elasticity problem herein. 
In addition, all other loss components obtained are lowest upon using the tanh activation as shown in Table \ref{T-1}.

Comparing the evolution of $\Delta_\mathcal{L}$, the convergence characteristics for the ReLU activation are better compared to the tanh with fewer fluctuations and rapid decrease in loss values as shown in Fig. \ref{Fig-4}-(a). 
However, the tanh illustrates better adaptability in the constitutive loss with an excellent convergence rate in  Fig. \ref{Fig-4}-(b). 
Out of the three activations, ReLU performs the worst possibly due to its derivative being discontinuous. 
However,  the total loss for all three activations is negligible  (loss value in the range below $10^{-4}$ to $10^{-5}$) within 200 epochs indicating the adaptability of the proposed PINNs framework to any of these activations provided the models are trained sufficiently long. 
In comparing the training time, the tanh activation takes longer for the same number of epochs compared to the other two. 
This coincides with the fact that the evolution of the total loss has a higher degree of discontinuity. 
However, the model with the ReLU activation trains the fastest possibly due to its linear nature. 
From the comparison, it can be concluded that although tanh is the best in terms of accuracy, however,  ReLU can be an optimal choice of activation considering both accuracy and training time for solving elasticity equation in the proposed PINNs framework. 
\begin{table}
	\centering
	\caption{Influence of different activation functions on the final values of various loss components (in $10^{-09}$) and training times $t_{tr}$ in the proposed PINNs model for solving linear elastic beam problem.}
	\begin{tabular}{c c c c c c c c c c c }
		\\[-0.5em]
		\hline
		\\[-0.8em] 
		Activation Function  & $\Delta^{{\Omega}}_\mathcal{L}$  & $\Delta^{c}_\mathcal{L}$  & $\Delta^{{\Gamma_u}}_\mathcal{L}$ & 
		$\Delta^{{\Gamma_t}}_\mathcal{L}$ & $\Delta^{{\fg u}}_\mathcal{L}$ & $\Delta^{{\fsg}}_\mathcal{L}$ & $\Delta^{{\fvep}}_\mathcal{L}$ &$\Delta_\mathcal{L}$ &\quad \vtop{\hbox{\strut $t_{tr}$}\hbox{\strut  ($min$)}}
		\\[-0.0em]
		\hline
		\\[-0.5em]
		ReLU & 107.16 & 43.43 & 14.51  & 36.75  & 24.97 & 1.07 & 5.48 & 233.37 & 9.4 
		\\
		\\[-0.5em]
		Sigmoid  & 30.96 & 54.33 & 517.38  & 126.14   & 37.85 & 124.51 & 592.82 & 1483.99 & 13.8
		\\
		\\[-0.5em]
		tanh  & 4.56 & 0.73 & 31.47  & 25.64  & 3.11 & 9.60 & 10.45 & 85.56 & 15.7
		\\
		\\[-0.5em]
		\hline
	\end{tabular}
\label{T-1}
\end{table}
\\
\\
{\bf 4.2.4 Influence of network complexity:}
\\
\\
It is worth mentioning that the PINNs approximations are sensitive to network architecture including the depth of the hidden layer and the number of network parameters. 
In this section,  the influence of network architecture parameters, i.e., the number of neurons in each hidden layer $\mathscr{N}$, and the number of hidden layers $L_n$ on the accuracy and the efficiency of the PINNs solution are explored. 
Since the tanh activation performs the best in terms of accuracy (see previous section), it is chosen as the activation for different networks used in the following experiments. 
\begin{table}
\centering
\caption{Influence of network parameters $\mathscr{N}$ and  $L_n$ on training times $t_{tr}$ and final values various loss components (in $10^{-09}$)  for tanh activation. }
\begin{tabular}{c c c c c c c c c c c }
	\\[-0.5em]
	\hline
	\\[-0.8em] 
	\vtop{\hbox{\strut Network  }\hbox{\strut identifier}} & \vtop{\hbox{\strut $n_p$  }\hbox{\strut }} & \vtop{\hbox{\strut $t_{tr}$ }\hbox{\strut  ($min$)}} & $\Delta^{{\Omega}}_\mathcal{L}$  & $\Delta^{c}_\mathcal{L}$  & $\Delta^{{\Gamma_u}}_\mathcal{L}$ & 
	$\Delta^{{\Gamma_t}}_\mathcal{L}$ & $\Delta^{{\fg u}}_\mathcal{L}$ & $\Delta^{{\fsg}}_\mathcal{L}$ & $\Delta^{{\fvep}}_\mathcal{L}$ & $\Delta_\mathcal{L}$
	\\[-0.0em]
	\hline
	\\[-0.5em]
	N-1 ($\mathscr{N}=20$, $L_n=5$)  & 22,706 & 15.7 &  4.56 & 0.73 & 31.47  & 25.64  & 3.11 & 9.60 & 10.45 & 85.56  
	\\
	\\[-0.5em]
	N-2 ($\mathscr{N}=40$, $L_n=5$) & 113,530 & 23.8 & 2.21 & 90.39 & 77.73  & 59.58 & 4.29 & 24.16 & 78.39 & 336.75  
	\\
	\\[-0.5em]
	N-3 ($\mathscr{N}=20$, $L_n=10$) & 54,494 & 18.3  & 6.89 & 0.89 & 12.73  & 65.42  & 13.01 & 17.19 & 4.67 & 120.8  
	\\
	\\[-0.5em]
	N-4 ($\mathscr{N}=40$, $L_n=10$) & 272,472 & 32.3  & 2.78 & 3.67 & 18.78  & 12.63  & 24.19 & 43.10 & 2.49 & 107.64 
	\\
	\\[-0.5em]
	\hline
\end{tabular}
\label{T-2}
\end{table}
\begin{figure}
\noindent
\centering
\includegraphics[width=0.95\linewidth]{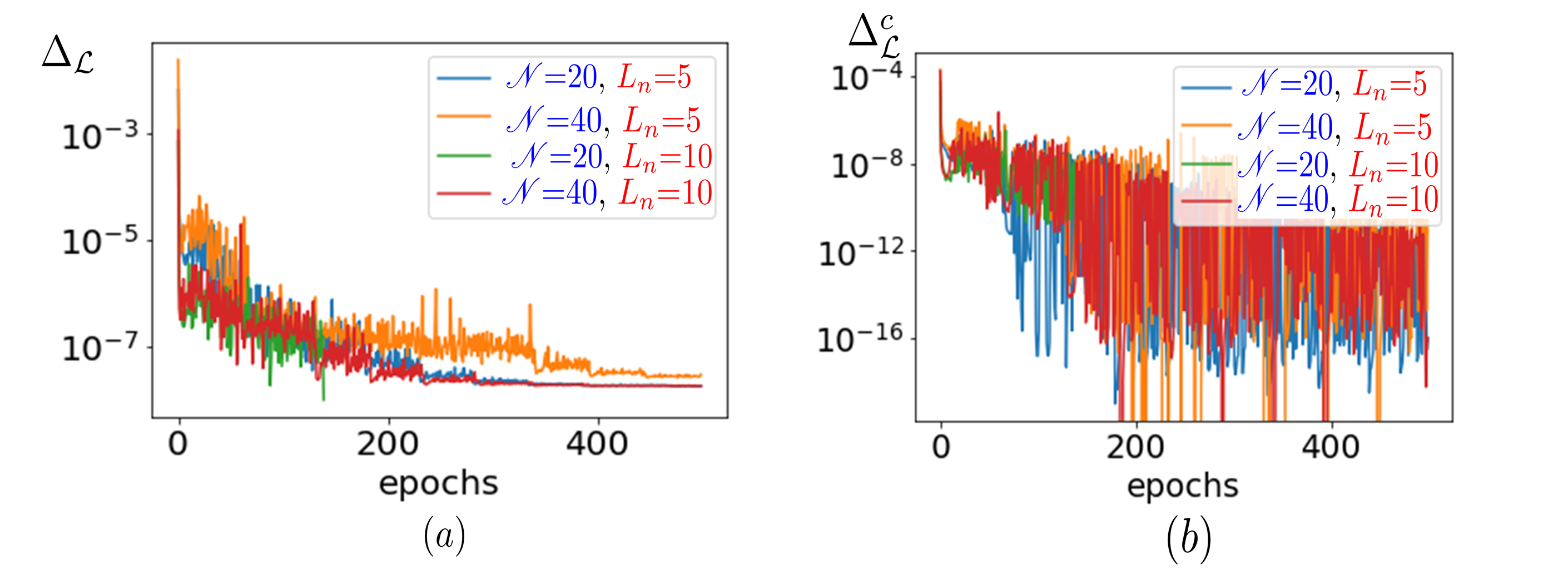}
\caption{\label{Fig-5} Comparison of (a) total loss $\Delta^{{\Omega}}_\mathcal{L}$; (b) constitutive loss  $\Delta^{{\Omega}}_\mathcal{L}$ for various combinations of network parameters $\mathscr{N}$ and  $L_n$ considering tanh activation function.}
\end{figure}

In the current study, four different networks considering the combinations $\mathscr{N}= 20, 40 $, and $L_n = 5, 10$ are tested, and values of different loss components at the end of the training, training duration ($t_{tr}$), along with model complexities in terms of network parameters ($n_p$) for these architectures are presented in Table. \ref{T-2}. 
For fair comparison, $N_c=5,000$ for all experiments. 
The evolution of the total loss $\Delta_\mathcal{L}$ and the constitutive loss  $\Delta^{{\Omega}}_\mathcal{L}$ for these networks are shown in Fig. \ref{Fig-5}. 
From the comparisons, for the chosen number of collocation points relatively shallow network $\mathscr{N}= 20$, $L_n = 5$ provides the best performance in terms of $\Delta_\mathcal{L}$ and $\Delta^{{\Omega}}_\mathcal{L}$ at the end of training. 
Additionally, the time required for training is faster due to a significantly lower number of network parameters. 
However, for a relatively deeper network, $\mathscr{N}= 20$, $L_n = 10$ with increased network complexity, the performance of the model degrades with respect to loss values as shown in Table. \ref{T-2} possibly due to an increase in variability and reduction in bias. 
Interestingly, an increase in the number of neurons $\mathscr{N}= 40$ while maintaining the depth of the network ($L_n = 5$) leads to the worst performance which can be attributed to over-fitting \citep{bilbao2017overfitting,jabbar2015methods}. 
The epoch evolution of the loss for various network architectures demonstrates the efficacy of a relatively shallow network 
with significantly faster training for solving elasticity problems in the proposed PINNs framework. 
\\
\\
{\bf 5. PINNs formulation for linear elastic plate theory :}
\\
\\
In this section, the PINNs framework is expanded for the solution of the classical Kirchhoff-Love thin plate \citep{Timoshenko-plate-1959} subjected to a transverse loading in linearly elastic plate theory. 
In the subsequent section, the Kirchhoff-Love theory has been briefly described; PINNs formulation for solving the governing fourth-order biharmonic partial differential equation (PDE) for the solution of the thin plate is elaborated. 
For a benchmark problem, the proposed PINNs approach is applied for the solution of a simply supported rectangular plate under a transverse sinusoidal loading condition. 
\\
\\
{\bf 5.1  Kirchhoff-Love thin plate theory :}
\\
\\
Thin plates are structurally planar elements that have small thickness relative to their in-plane dimensions which can be simplified as a two-dimensional plate problem. 
According to the Kirchhoff-Love theory, the kinetics of a thin plate under the effect of a distributed transverse loading  $q=q(x,y)$ can be described by a fourth-order differential equation \citep{Timoshenko-plate-1959,Reddy-plate-2006}. 
\begin{equation}
\mathbf{\Delta}(\mathscr{D}\mathbf{\Delta} w)=q
\label{E-35}
\end{equation}
When the elastic plate is bounded in the domain $\Omega  \subset \mathbb{R}^2$, Eq. \ref{E-35} is known as the Kirchhoff-Love equation. 
In Cartesian coordinates, $w=w(x,y)$ represents the transverse displacement field, $\mathscr{D}=\mathscr{D} (x,y)$ is the bending stiffness of the plate, and $\mathbf{\Delta}=\partial^2/\partial x^2 + \partial^2/\partial y^2$ is the Laplace operator. 
Considering a homogeneous and isotropic plate (i.e., $\mathscr{D} \equiv$ constant ), Eq. \ref{E-35} becomes the biharmonic equation \citep{Timoshenko-plate-1959,Szilard-plate-1974}
\begin{equation}
\mathscr{D}	\mathbf{\Delta}^2 w= \mathscr{D} \left(\frac{\partial^4 w}{\partial x^4}+ 	2\frac{\partial^4 w}{\partial x^2 \partial y^2} + \frac{\partial^4 w}{\partial y^4}\right)=q
\label{E-36}
\end{equation}
Under appropriate boundary conditions, and with $\mathscr{D} (x,y)>0$ and $q(x,y) \ge 0$, both being known, the problem possesses a unique solution for the displacement $w(x,y)$. 
The set of solution variables includes the primitive variable deflection ${w}$, and the derived quantities, moments ${M}_{xx}$, ${M}_{yy}$, ${M}_{xy}=-{M}_{yx}$, and shearing forces ${Q}_{xx}$, ${Q}_{yy}$. 
The expressions for the derived fields are, 
\bey
M_{xx}=-\mathscr{D}\left(\frac{\partial^2 w}{\partial x^2}+\nu \frac{\partial^2 w}{\partial y^2}\right); \quad M_{yy}=-\mathscr{D}\left(\frac{\partial^2 w}{\partial y^2}+\nu \frac{\partial^2 w}{\partial x^2}\right); \quad {M}_{xy}=-\mathscr{D}(1-\nu)\left( \frac{\partial^2 w}{\partial x \partial y}\right)
\label{E-37}
\eey
\bey
Q_{xx}=\frac{\partial M_{yx}}{\partial y}+
\frac{\partial M_{xx}}{\partial x}=
-\mathscr{D} \frac{\partial}{\partial x}\left(\frac{\partial^2 w}{\partial x^2}+ \frac{\partial^2 w}{\partial y^2}\right) ; \quad Q_{yy}=\frac{\partial M_{yy}}{\partial y}-
\frac{\partial M_{xy}}{\partial x}=
-\mathscr{D} \frac{\partial}{\partial y}\left(\frac{\partial^2 w}{\partial x^2}+ \frac{\partial^2 w}{\partial y^2}\right)
\label{E-38}
\eey
\\
\\
\begin{figure}
\noindent
\centering
\includegraphics[width=0.7\linewidth]{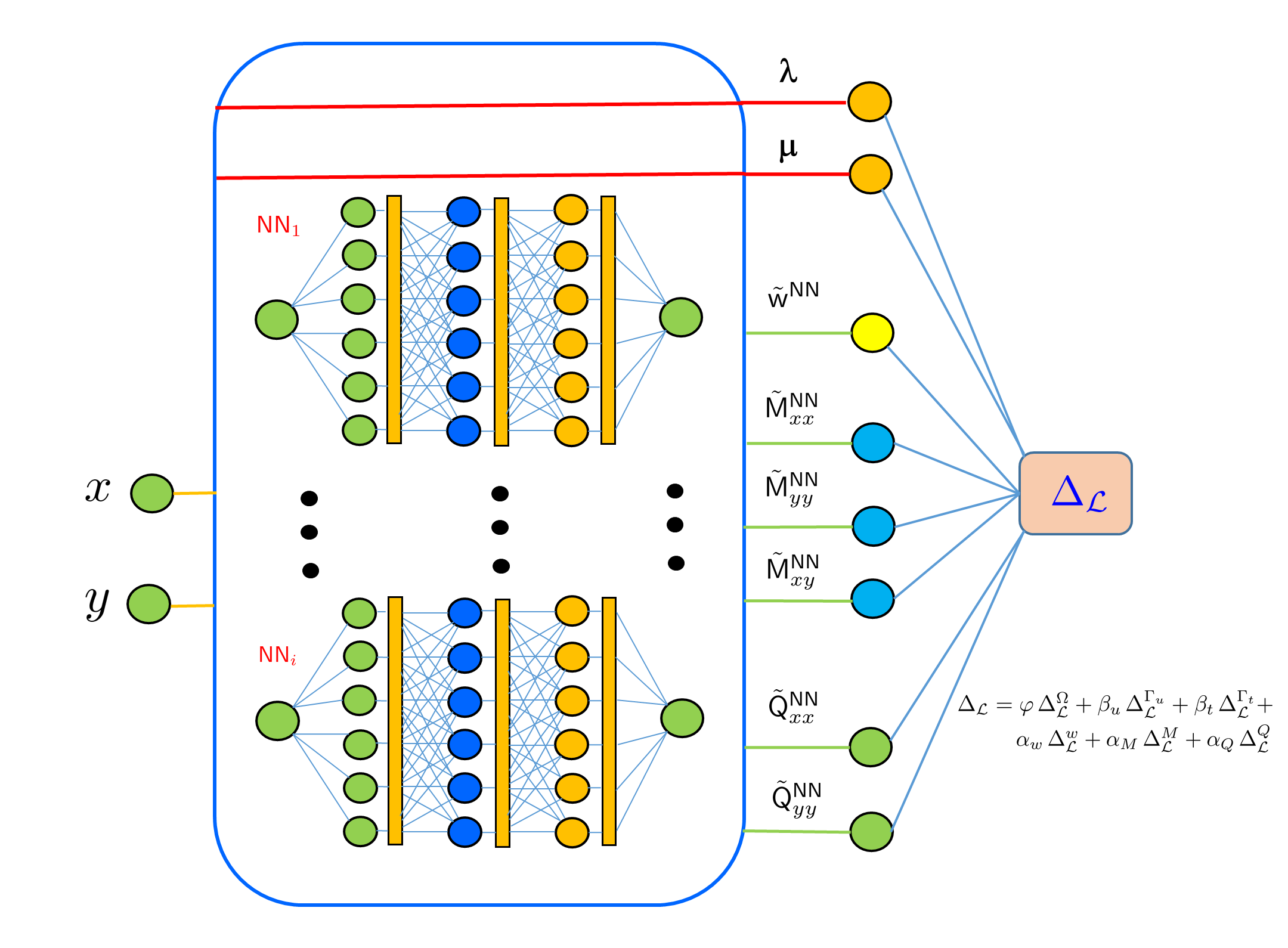}
\caption{\label{Fig-7} PINNs network architecture for  solving Kirchhoff-Love thin plate problem governed by  biharmonic equation consisting  of multi-ANN ($\mathsf{NN}_i \,\forall\, i=1,k$) for each field variables $\tilde{\mathsf{w}}^{\mathsf{NN}}(\fg{x})$, $\tilde{\mathsf{M}}^{\mathsf{NN}}_{xx}(\fg{x})$, $\tilde{\mathsf{M}}^{\mathsf{NN}}_{xy}(\fg{x})$,
$\tilde{\mathsf{M}}^{\mathsf{NN}}_{yy}(\fg{x})$,
$\tilde{\mathsf{Q}}^{\mathsf{NN}}_{xx}(\fg{x})$,
and $\tilde{\mathsf{Q}}^{\mathsf{NN}}_{yy}(\fg{x})$ with independent variable  $\fg{x}=(x, y) $ as input features.}
\end{figure}

{\bf 5.2 PINNs formulation for the Biharmonic equation:}
\\
\\
For solving the Biharmonic equation using the PINNs framework, the input features are the spatial coordinates $\fg{x} := (x, y)$; the field variables, $w(\fg{x})$, $\fg{M}(\fg{x})$, and $\fg{Q}(\fg{x})$ are obtained using multiple densely connected independent ANNs,
with each network approximating one of the outputs (Fig. \ref{Fig-7}). 
Different field variables approximated by the NNs are as follows:
\begin{equation}
w(\fg{x})  \simeq \Xi^{\mathsf{NN}}_{w} =
\tilde{\mathsf{w}}^{\mathsf{NN}}(\fg{x})  
\label{E-39}
\end{equation}
\begin{equation}
\fg{M}(\fg{x})  \simeq \Xi^{\mathsf{NN}}_{{M}} =
\begin{bmatrix}
\tilde{\mathsf{M}}^{\mathsf{NN}}_{xx}(\fg{x}) & \tilde{\mathsf{M}}^{\mathsf{NN}}_{xy}(\fg{x}) \\
\tilde{\mathsf{M}}^{\mathsf{NN}}_{yx}(\fg{x}) & \tilde{\mathsf{M}}^{\mathsf{NN}}_{yy}(\fg{x})
\end{bmatrix};\quad\quad \fg{Q}(\fg{x})  \simeq \Xi^{\mathsf{NN}}_{\fg{Q}} = 
\begin{bmatrix}
\tilde{\mathsf{Q}}^{\mathsf{NN}}_{xx}(\fg{x})  \\
\tilde{\mathsf{Q}}^{\mathsf{NN}}_{yx}(\fg{x})
\end{bmatrix}
\label{E-40}
\end{equation}
\noindent where, $\Xi^{\mathsf{NN}}_{w}$, $\Xi^{\mathsf{NN}}_{\fg{M}}$, and $\Xi^{\mathsf{NN}}_{\fg{Q}}$ are the neural network appoximations. 
From the NN approximations of the fields, the muti-objective loss function $\Delta_\mathcal{L} (\fg{x}, \theta)$ can be defined as:
\begin{equation}
\Delta_\mathcal{L} (\fg{x}, \theta)=\varphi\, \Delta^{{\Omega}}_\mathcal{L}+
\beta_u\,\Delta^{{\Gamma_u}}_\mathcal{L}+
\beta_t\,\Delta^{{\Gamma_t}}_\mathcal{L}+
\alpha_{w}\, \Delta^{w}_\mathcal{L}+ \alpha_{{ M}}\, \Delta^{{ M}}_\mathcal{L}+ \alpha_{{ Q}}\, \Delta^{{ Q}}_\mathcal{L}
\label{E-41}
\end{equation}
\noindent where, $\Delta^{{\Omega}}_\mathcal{L}$, $\Delta^{{\Gamma_u}}_\mathcal{L}$, $\Delta^{{\Gamma_t}}_\mathcal{L}$  are the losses  in the domain $\Omega$, and along the boundaries $\Gamma_u$ and $\Gamma_t$, respectively. 
Their expressions are, 
\bey
\Delta^{{\Omega}}_\mathcal{L} &=& \frac{1}{N_c^{\Omega}} \sum_{l=1}^{N_c^{\Omega}} \lVert \mathbf{\nabla^2} \mathbf{\nabla^2} w -\frac{\hat{q}}{\mathscr{D}} \rVert 
\label{E-42}\\
\Delta^{{\Gamma_u}}_\mathcal{L}&=& \frac{1}{N_c^{\Gamma_u}} \sum_{k=1}^{N_c^{\Gamma_u}} \lVert \mathbf{\Xi^{\mathsf{NN}}_{{w}} (\fg{x}}_{k \vert \Gamma_u})  -{\bar{w}}(\fg{x}_{k \vert \Gamma_u})| \rVert
\label{E-43}\\
\Delta^{{\Gamma_t}}_\mathcal{L}&=& \frac{1}{N_c^{\Gamma_t}} \sum_{j=1}^{N_c^{\Gamma_t}} \lVert \mathbf{\Xi^{\mathsf{NN}}_{M} (\fg{x}}_{j \vert \Gamma_t})  -\fg{{\bar{M}}}(\fg{x}_{j \vert \Gamma_t})| \rVert
\label{E-44}
\eey
\noindent where, $\left\{\fg{x}_{1 \vert \Omega}, ..., \fg{x}_{N_c^{\Omega} \vert \Omega}\right\}$,
$\left\{\fg{x}_{1 \vert {\Gamma_u}}, ..., \fg{x}_{N_c^{{\Gamma_u}} \vert {\Gamma_u}}\right\}$,
$\left\{\fg{x}_{1 \vert {\Gamma_t}}, ..., \fg{x}_{N_c^{{\Gamma_t}} \vert {\Gamma_t}}\right\}$
are the collocation points over the domain $\Omega$­, and along the boundaries $\Gamma_u$ and $\Gamma_t$, respectively; 
$\varphi \in  \mathbb{R}^+$ is the penalty coefficient for imposing the biharmonic relation in Eq. \ref{E-36}. 
Additionally, data driven estimates of $w(\fg{x})$, $\fg{M}(\fg{x})$, and $\fg{Q}(\fg{x})$ at the collocation points across $\Omega$ are used to define $\Delta_\mathcal{L} (\fg{x}, \theta)$. 
\bey
\Delta^{w}_\mathcal{L}&=& \frac{1}{N_c^{\Omega}} \sum_{l=1}^{N_c^{\Omega}} \lVert \mathbf{ \Xi^{\mathsf{NN}}_{{w}} (\fg{x}}_{l \vert \Omega})  -{\hat{w}}(\fg{x}_{l \vert \Omega}) \rVert 
\label{E-45}\\
\Delta^{{ M}}_\mathcal{L}&=&\frac{1}{N_c^{\Omega}} \sum_{l=1}^{N_c^{\Omega}} \lVert \mathbf{ \Xi^{\mathsf{NN}}_{M} (\fg{x}}_{l \vert \Omega})  -\fg{\hat{M}}(\fg{x}_{l \vert \Omega}) \rVert
\label{E-46}\\
\Delta^{{ Q}}_\mathcal{L}&=& \frac{1}{N_c^{\Omega}} \sum_{l=1}^{N_c^{\Omega}} \lVert \mathbf{ \Xi^{\mathsf{NN}}_{\mathbf{Q}} (\fg{x}}_{l \vert \Omega})  -\fg{\hat{Q}}(\fg{x}_{l \vert \Omega}) \rVert
\label{E-47}
\eey
Here, ${\hat{w}}(\fg{x}_{l \vert \Omega})$, $\fg{\hat{M}}(\fg{x}_{l \vert \Omega})$, and $\fg{\hat{Q}}(\fg{x}_{l \vert \Omega})$ are obtained by means of analytical or high-fidelity numerical solutions. 
Note, $\alpha_i =1 ; \,\,\forall\,\, i=w, M, Q$ for data-driven enhancement coupled with physics-informed regression by forcing the PDE constraints in Eqs. \ref{E-36}-\ref{E-38}. 
Whereas, $\alpha_i = 0$ switches off the data-driven enhancement of accuracy of the NN approximations. 
The loss function in Eq. \ref{E-41} can either be used for obtaining PINNs approximations of  $w(\fg{x})$, $\fg{M}(\fg{x})$, and $\fg{Q}(\fg{x})$ (i.e., forward problem ), or identification of model parameters $\lambda$ and $\mu$ (i.e., inverse problem ). 
\\
\\
{\bf 5.3 Simply supported Kirchhoff-Love plate:}
\\
\\
A simply supported rectangular plate of size $(a \times b)$ under a sinusoidal load $q (x,y)=q_0 \sin \frac{\pi x}{a} \sin \frac{\pi y}{b}$ is considered in Cartesian coordinates as shown in Fig. \ref{Fig-7}. 
Here, $q_0$ is the intensity of the load at the center of the plate.	
\begin{figure}
\noindent
\centering
\includegraphics[width=1\linewidth]{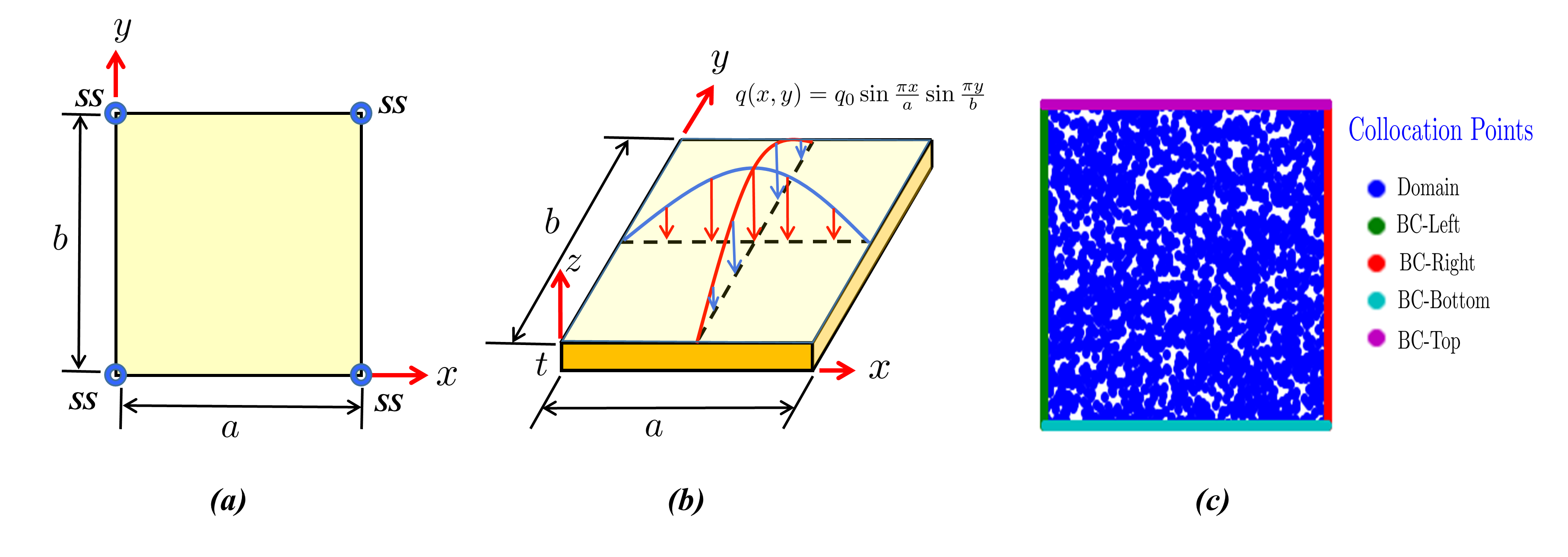}
\caption{\label{Fig-7}  
Benchmark problem setup for Kirchhoff-Love plate:  (a, b) simply supported rectangular plate of  $a=200$ cm and $b=300$ cm with thickness $t=1$ cm subjected to transverse sinusoidal loading of intensity $q_0= 9.806 \times 10^{-4}$ MPa;   (b) distributions of total collocations points $N_c=10,000$  on the problem domain and various boundaries during PINNs training.}
\end{figure}
The following boundary conditions are applied at the simply supported (SS) edges: 
\bey
w &=& 0; \quad\quad\quad \frac{\partial^2 w}{\partial x^2} =0 \quad\quad \mbox{for}\,\, x=0\,\, \mbox{and}\,\, x=a
\label{E-48}\\
w &=& 0; \quad\quad\quad \frac{\partial^2 w}{\partial y^2} =0 \quad\quad \mbox{for}\,\, y=0\,\, \mbox{and}\,\, y=b
\label{E-49}
\eey

{\bf 5.3.1  Analytical solution:} Along with the governing equation in Eq. \ref{E-36} and the boundary conditions in Eqs. \ref{E-48}- \ref{E-49}, the analytical solutions of $w$ are obtained as: 		
\begin{equation}
w= \frac{q_0}{\pi^4(\frac{1}{a^2}+\frac{1}{b^2})^2} \sin \frac{\pi x}{a} \sin \frac{\pi y}{b} 
\label{E-50}
\end{equation}
Utilizing Eqs. \ref{E-37}-\ref{E-38}, analytical solutions for the moments $M_{xx}$, $M_{yy}$, $M_{xy}$ and the shearing forces, $Q_{xx}$, $Q_{yy}$ are obtained as: 
\bey
M_{xx} &=& \frac{q_0}{\pi^2\left(\frac{1}{a^2}+\frac{1}{b^2}\right)^2}\left(\frac{1}{a^2}+\frac{\nu}{b^2}\right) \sin \frac{\pi x}{a} \sin \frac{\pi y}{b}  
\label{E-51}\\
M_{yy} &=& \frac{q_0}{\pi^2\left(\frac{1}{a^2}+\frac{1}{b^2}\right)^2}\left(\frac{\nu}{a^2}+\frac{1}{b^2}\right) \sin \frac{\pi x}{a} \sin \frac{\pi y}{b}  
\label{E-52}\\
M_{xy} &=& \frac{q_0(1-\nu)}{\pi^2\left(\frac{1}{a^2}+\frac{1}{b^2}\right)^2 ab}\left(\frac{\nu}{a^2}+\frac{1}{b^2}\right) \cos \frac{\pi x}{a} \cos \frac{\pi y}{b}  
\label{E-53}\\
Q_{xx} &=& \frac{q_0}{\pi a\left(\frac{1}{a^2}+\frac{1}{b^2}\right)} \cos \frac{\pi x}{a} \sin \frac{\pi y}{b}  
\label{E-54}\\
Q_{yy} &=& \frac{q_0}{\pi a\left(\frac{1}{a^2}+\frac{1}{b^2}\right)} \sin \frac{\pi x}{a} \sin \frac{\pi y}{b}  
\label{E-55}
\eey
These analytical solutions, $w(\fg{x})$, $\fg{M}(\fg{x})$, and $\fg{Q}(\fg{x})$ have been utilized as 
${\hat{w}}(\fg{x}_{l \vert \Omega})$, $\fg{\hat{M}}(\fg{x}_{l \vert \Omega})$, and $\fg{\hat{Q}}(\fg{x}_{l \vert \Omega})$ for data driven enhancement in Eqs. \ref{E-45}-\ref{E-47}, respectively for the PINNs approximations of the field variables.
\\
\\
{\bf 5.4  PINNs solutions for the Biharmonic equation:}	
\\
\\
For the benchmark problem, a rectangular plate ($a=200$ cm, $b=300$ cm) with thickness $t=1$ cm is considered with the following material properties: Young's modulus of elasticity $E$= 202017.03 MPa, Poisson's ratio $\nu =0.25$, and flexural rigidity $\mathscr{D}$= 17957 N-m. 
The sinusoidal load intensity $q_0= 9.806 \times 10^{-4}$ MPa is presribed as shown in Fig. \ref{Fig-7}.  A similar problem has been also solved in the recent work\citep{Haghighat-2021-Biharmonic}.
Unless otherwise stated, the total number of randomly distributed collocation points, $N_c= 10,000$ is used during the training of the PINNs model. 
Additionally, a learning rate of 0.001, and a batch size of 50 were prescribed for optimal accuracy and faster convergence of the optimization scheme. 
For better accuracy during training, the Adam optimization scheme is employed with 1000 epochs. 
In the present study, three different activation functions were tested (see section 5.4.1). 
\begin{figure}
\noindent
\centering
\includegraphics[width=1.1\linewidth]{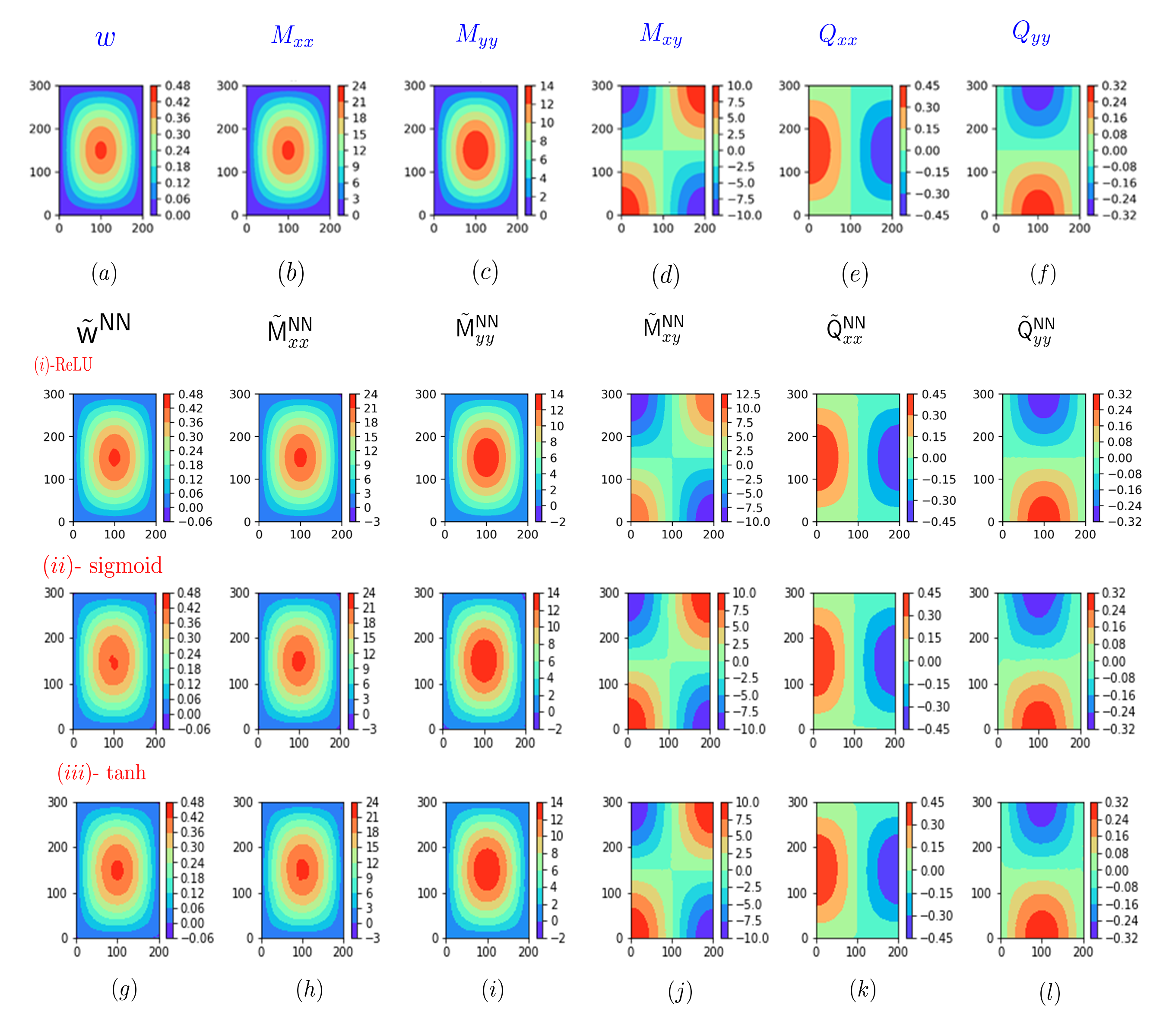}
\caption{\label{Fig-8} Solution of field variables obtained from (a-f) analytical solutions (left to right): $w$, $M_{xx}$, $M_{yy}$, $M_{xy}$, $Q_{xx}$, and $Q_{yy}$  ; (g-l) proposed PINNs results  (left to right): $\tilde{\mathsf{w}}^{\mathsf{NN}}$, $\tilde{\mathsf{M}}^{\mathsf{NN}}_{xx}$, $\tilde{\mathsf{M}}^{\mathsf{NN}}_{xy}$, $\tilde{\mathsf{M}}^{\mathsf{NN}}_{yy}$, $\tilde{\mathsf{Q}}^{\mathsf{NN}}_{xx}$, and  $\tilde{\mathsf{Q}}^{\mathsf{NN}}_{yy}$   for activation functions (i) ReLU, (ii) sigmoid, and (iii) tanh.  } 
\end{figure}

\begin{figure}
\noindent
\centering
\includegraphics[width=1\linewidth]{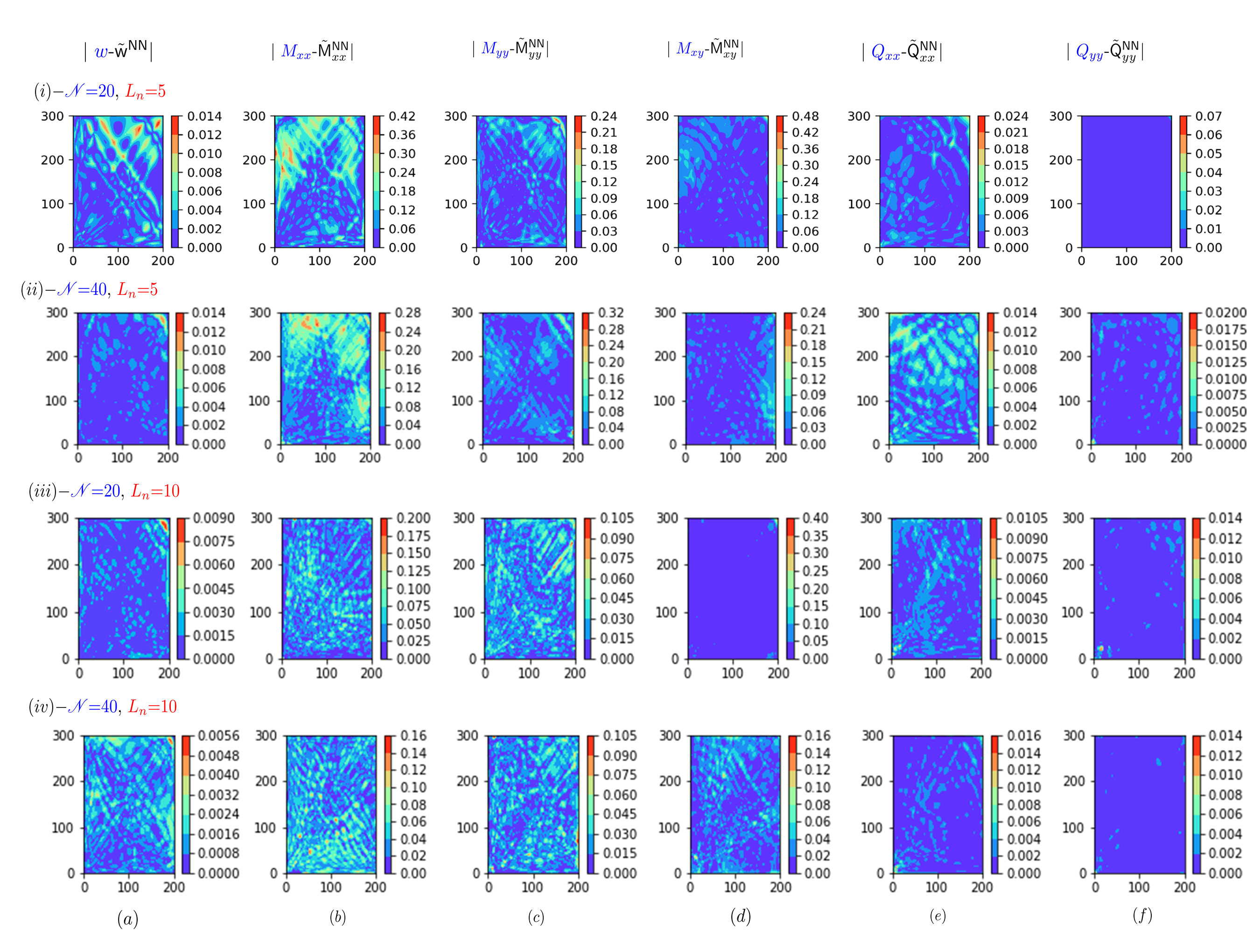}
\caption{\label{Fig-11} Absolute  error of field variables between analytical solution and  PINNs results (a) $\vert w-\tilde{\mathsf{w}}^{\mathsf{NN}} \vert$; (b) $\vert M_{xx}-\tilde{\mathsf{M}}^{\mathsf{NN}}_{xx} \vert$; (c) $\vert M_{yy}-\tilde{\mathsf{M}}^{\mathsf{NN}}_{yy} \vert $; (d)  $\vert M_{xy}-\tilde{\mathsf{M}}^{\mathsf{NN}}_{xy} \vert$;
(e) $\vert Q_{xx}-\tilde{\mathsf{Q}}^{\mathsf{NN}}_{xx} \vert $; and  (f) $\vert Q_{yy}-\tilde{\mathsf{Q}}^{\mathsf{NN}}_{yy} \vert $    for  various network parameters  (i) $\mathscr{N}=20, L_n=5$,  (ii) $\mathscr{N}=40, L_n=5$,   (iii) $\mathscr{N}=20, L_n=10$, and (iv)  $\mathscr{N}=40, L_n= 10$.}
\end{figure}

\begin{figure}
\noindent
\centering
\includegraphics[width=0.95\linewidth]{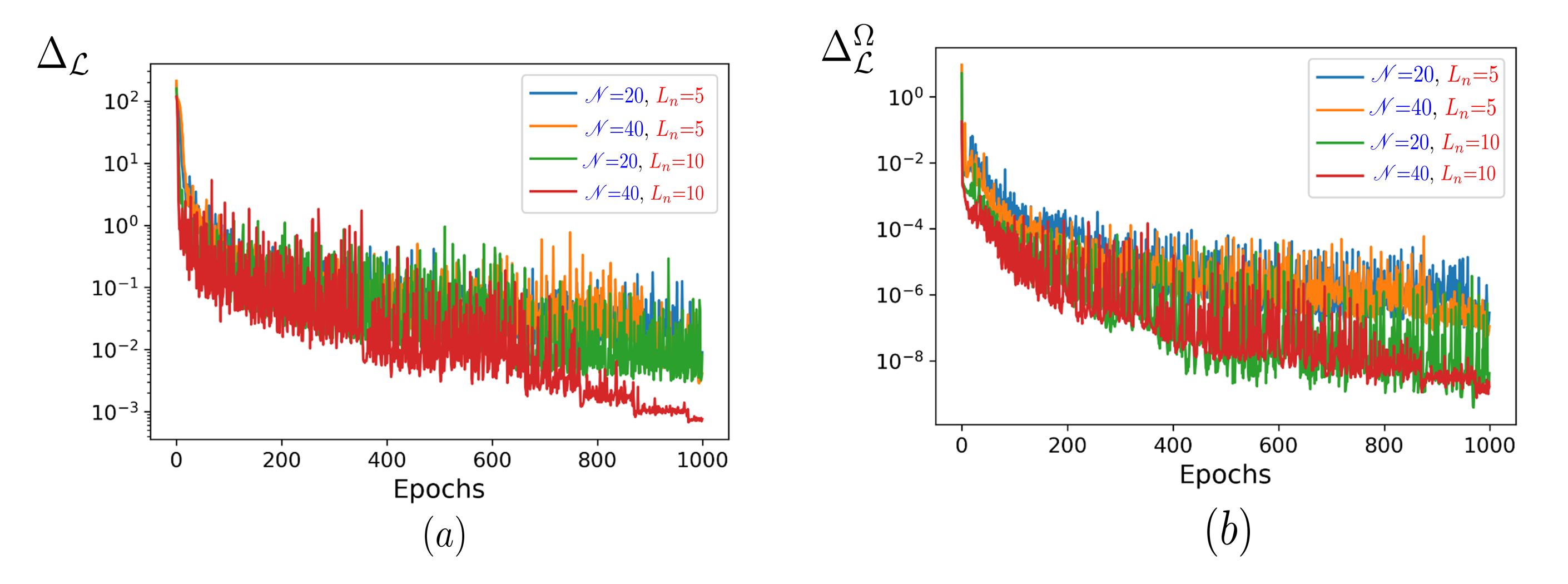}
\caption{\label{Fig-12} Comparison of (a) total loss $\Delta^{{\Omega}}_\mathcal{L}$; (b) constitutive loss  $\Delta^{{\Omega}}_\mathcal{L}$ for various combinations of network parameters $\mathscr{N}$ and  $L_n$ considering  ReLU activation.}
\end{figure}

In Fig. \ref{Fig-8}(a--f), the analytical solution for various fields including plate deflection $w$, moments $M_{xx}$, $M_{yy}$, $M_{xy}$, and shearing  forces $Q_{xx}$, and $Q_{yy}$ in Eqs. \ref{E-50}-\ref{E-55} are shown. 
Corresponding approximations from PINNs for various activation functions are shown in  Fig. \ref{Fig-8} (a--f) which illustrate the efficacy of the proposed model in terms of accuracy and robustness as excellent agreement with the analytical solutions is evident. 
\begin{figure}
\noindent
\centering9
\includegraphics[width=1\linewidth]{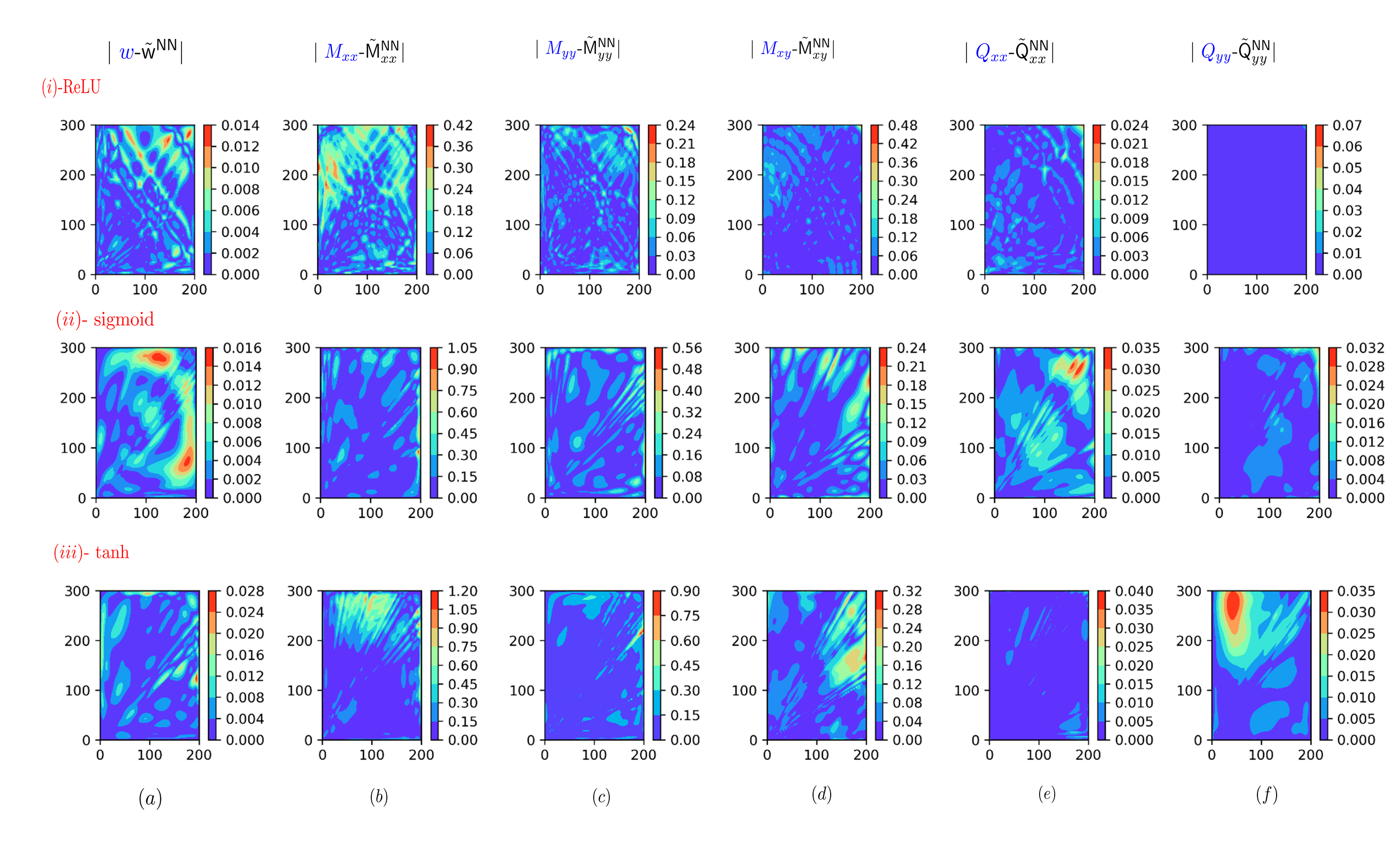}
\caption{\label{Fig-9} Absolute  error of field variables between analytical solution and  PINNs results (a) $\vert w-\tilde{\mathsf{w}}^{\mathsf{NN}} \vert$; (b) $\vert M_{xx}-\tilde{\mathsf{M}}^{\mathsf{NN}}_{xx} \vert$; (c) $\vert M_{yy}-\tilde{\mathsf{M}}^{\mathsf{NN}}_{yy} \vert $; (d)  $\vert M_{xy}-\tilde{\mathsf{M}}^{\mathsf{NN}}_{xy} \vert$;
(e) $\vert Q_{xx}-\tilde{\mathsf{Q}}^{\mathsf{NN}}_{xx} \vert $; and  (f) $\vert Q_{yy}-\tilde{\mathsf{Q}}^{\mathsf{NN}}_{yy} \vert $    for activation functions (i) ReLU, (ii) sigmoid, and (iii) tanh. }
\end{figure}
\begin{figure}
\noindent
\centering
\includegraphics[width=0.95\linewidth]{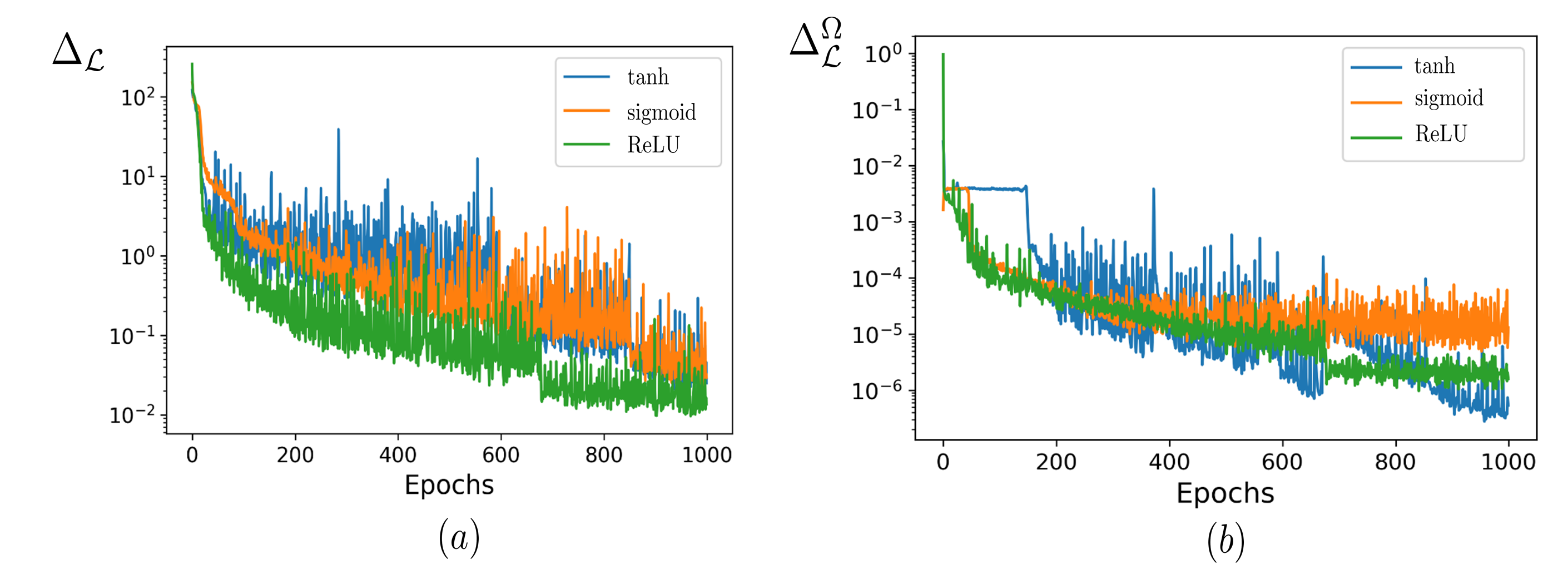}
\caption{\label{Fig-10} Comparison of (a) total loss $\Delta_\mathcal{L}$; (b) constitutive loss  $\Delta^{{\Omega}}_\mathcal{L}$ during training  for  tanh, sigmoid and ReLU activation functions for  network parameters  $\mathscr{N}=20, L_n=5$.}
\end{figure}
\\
\\
{\bf 5.4.1  Influence of the activation function:}
\\
\\	
The accuracy of the field variables and epoch evolution of the loss functions are explored for various activation functions for solving the fourth-order biharmonic PDE. 
To this end, three different activations, i.e., ReLU, sigmoid, and tanh are selected; the network used is defined by $\mathscr{N}=20, L_n=5$. 
The corresponding results are depicted in Fig. \ref{Fig-8} (g--l). 
Based on the results, all the activations perform well as the NN approximations are in good agreement with the analytical solutions both qualitatively and quantitatively. 
For further insight into the influence of an activation function on the accuracy of the solutions, the absolute error between the analytical solutions and the PINNs approximations for each field variable is compared for the solutions obtained with different activations in Fig. \ref{Fig-9} (a--f). 
From the comparison, ReLU provides the least absolute error distributions in solving the Biharmonic equation for the simply supported plate. 
Although, the sigmoid activation provides the best result for $\vert M_{xy}-\tilde{\mathsf{M}}^{\mathsf{NN}}_{xy} \vert$, the absolute error for the rest of the fields is higher compared to the solutions obtained with ReLU. 
Because of the sinusoidal nature of the solution, it was expected that tanh activation might be specifically suitable for this problem.  Surprisingly, tanh provides worse results compared to ReLU and sigmoid activations. 
This can be due to the complex nature of the solution space, where ReLU can provide better adaptability during training. 
\begin{table}
\centering
\caption{Influence of different activation functions on the  final values of  various loss components (in $10^{-05}$) and training times $t_{tr}$ in  the proposed PINNs model for solving biharmonic PDE.}
\begin{tabular}{c c c c c  c c c c }
\\[-0.5em]
\hline
\\[-0.8em] 
Activation Function  & $\Delta^{{\Omega}}_\mathcal{L}$  & $\Delta^{{\Gamma_t}}_\mathcal{L}$ & $\Delta^{{\Gamma_u}}_\mathcal{L}$ &$\Delta^{w}_\mathcal{L}$ & $\Delta^{{ M}}_\mathcal{L}$ & $\Delta^{{Q}}_\mathcal{L}$ & $\Delta_\mathcal{L}$ & \vtop{\hbox{\strut $t_{tr}$ }\hbox{\strut  ($min$)}}
\\[-0.0em]
\hline
\\[-0.5em]
ReLU & 5.34  & 132.31 & 1672.91  & 278.43  & 498.76 & 101.36 & 2689.11 & 23.1
\\
\\[-0.5em]
Sigmoid  & 63.07 & 980.67 & 4601.60  & 1707.50 & 987.89 & 117.56 & 8458.29 & 25.8
\\
\\[-0.5em]
tanh  & 0.12 & 7138.43 & 9807.31  & 6809.34  & 397.89 & 500.37 & 24653.46 & 34.6 
\\
\\[-0.5em]
\hline
\end{tabular}
\label{T-3}
\end{table} 
Furthermore, in Fig. \ref{Fig-10}, the epoch evolution of the total loss $\Delta^{{\Omega}}_\mathcal{L}$, and constitutive loss $\Delta^{{\Omega}}_\mathcal{L}$ is compared for different activation functions. 
For a particular epoch, ReLU performs better than the other two activations for $\Delta_\mathcal{L}$. 
For  $\Delta^{{\Omega}}_\mathcal{L}$, tanh activation shows better convergence and the lowest loss value at the end of training due to the sinusoidal nature of the solution of the Biharmonic PDE. 
However, the fluctuations in the loss curve for tanh have a relatively higher variance compared to ReLU and sigmoid. 
As reported in Table \ref{T-3}, overall, performance in terms of various loss components at the end of training is superior for the ReLU activation for solving the Biharmonic PDE using the proposed PINNs framework. 
Additionally, the model with the ReLU activation requires the least training time $t_{tr}$, indicating better convergence and faster computation of the forward and backpropagation steps. 
\\
\\
{\bf 5.4.2  Influence of network parameters:}
\\
\\	
As was found for the linear elasticity problem, PINNs solutions are sensitive to the NN architecture. 
Various parameters that influence the NN architectures, the number of neurons in each hidden layer $\mathscr{N}$, and the total number of hidden layers $L_n$, on the accuracy of the model and the efficiency of training the model have been explored herein. 
Because of its superior performance for the problem, ReLU is chosen as the activation function. 
Four different networks with combinations $\mathscr{N}= 20, 40$, and  $L_n = 5, 10$ were trained. 
Corresponding network parameters ($n_p$), model training time ($t_{tr}$), and values of different loss components at the end of training have been presented in Table. \ref{T-4}. 
The comparisons of the absolute error between the analytical solutions and the PINNs approximations for each field are shown in Fig. \ref{Fig-11}. 
Comparisons of the total loss $\Delta_\mathcal{L}$, the constitutive loss $\Delta^{{\Omega}}_\mathcal{L}$ for various combinations of network parameters, $\mathscr{N}$ and  $L_n$ are shown in Fig. \ref{Fig-12}.

Based on the comparisons shown in Fig. \ref{Fig-11}, increased network depth improves the accuracy of the PINNs approximations for all variables. 
Predictions by both networks with $L_n = 10$ are superior compared to the analytical solutions for the chosen number of collocation points. 
On the other hand, an increase in the number of neurons in each layer increases model prediction variance which is reflected in the higher absolute error comparisons for $\mathscr{N}= 20, 40$ and $L_n = 10$. 
Similar conclusions may be drawn based on Fig. \ref{Fig-12} and Table. \ref{T-4}. 
The total and constitutive losses are minimum for $\mathscr{N}= 40$ and $L_n = 10$ at the end of training. 
However, the approximations by this model have higher variance. 
Expectedly, more complex models (higher $L_n$), or with larger $n_p$, require longer training time $t_{tr}$. 
For the chosen number of collocation points, $L_n =10$ is optimal. 
\begin{table}
\centering
\caption{ Influence of network parameters $\mathscr{N}$ and  $L_n$ on training times $t_{tr}$ and final values of  various loss components (in $10^{-05}$)  for tanh activation.}
\begin{tabular}{c c c c c c c c c c c }
\\[-0.5em]
\hline
\\[-0.8em] 
\vtop{\hbox{\strut Network  }\hbox{\strut identifier}} & \vtop{\hbox{\strut $n_p$  }\hbox{\strut }} & \vtop{\hbox{\strut $t_{tr}$ }\hbox{\strut  ($min$)}} & $\Delta^{{\Omega}}_\mathcal{L}$   & $\Delta^{{\Gamma_u}}_\mathcal{L}$ & 
$\Delta^{{\Gamma_t}}_\mathcal{L}$ & $\Delta^{{w}}_\mathcal{L}$ & $\Delta^{{M}}_\mathcal{L}$ & $\Delta^{{Q}}_\mathcal{L}$ & $\Delta_\mathcal{L}$
\\[-0.0em]
\hline
\\[-0.5em]
N-1 ($\mathscr{N}=20$, $L_n=5$)  & 12,940 & 23.1 & 5.34  & 132.31 & 1672.91  & 278.43  & 498.76 & 101.36 & 2689.11 
\\
\\[-0.5em]
N-2 ($\mathscr{N}=40$, $L_n=5$) & 52,760 & 29.8 & 0.47 & 35.13 & 467.34 & 128.38  & 198.11 & 40.29 & 869.72   
\\
\\[-0.5em]
N-3 ($\mathscr{N}=20$, $L_n=10$) & 32,056 & 31.7  & 0.07 & 82.15 & 86.84  & 77.82  & 298.01 & 10.17 & 555.06  
\\
\\[-0.5em]
N-4 ($\mathscr{N}=40$, $L_n=10$) & 126,224 & 42.8  & 0.009 & 0.67 & 5.12  & 4.21  & 0.53 & 0.17 & 10.709
\\
\\[-0.5em]
\hline
\end{tabular}
\label{T-4}
\end{table}
\begin{figure}
\noindent
\centering
\includegraphics[width=1\linewidth]{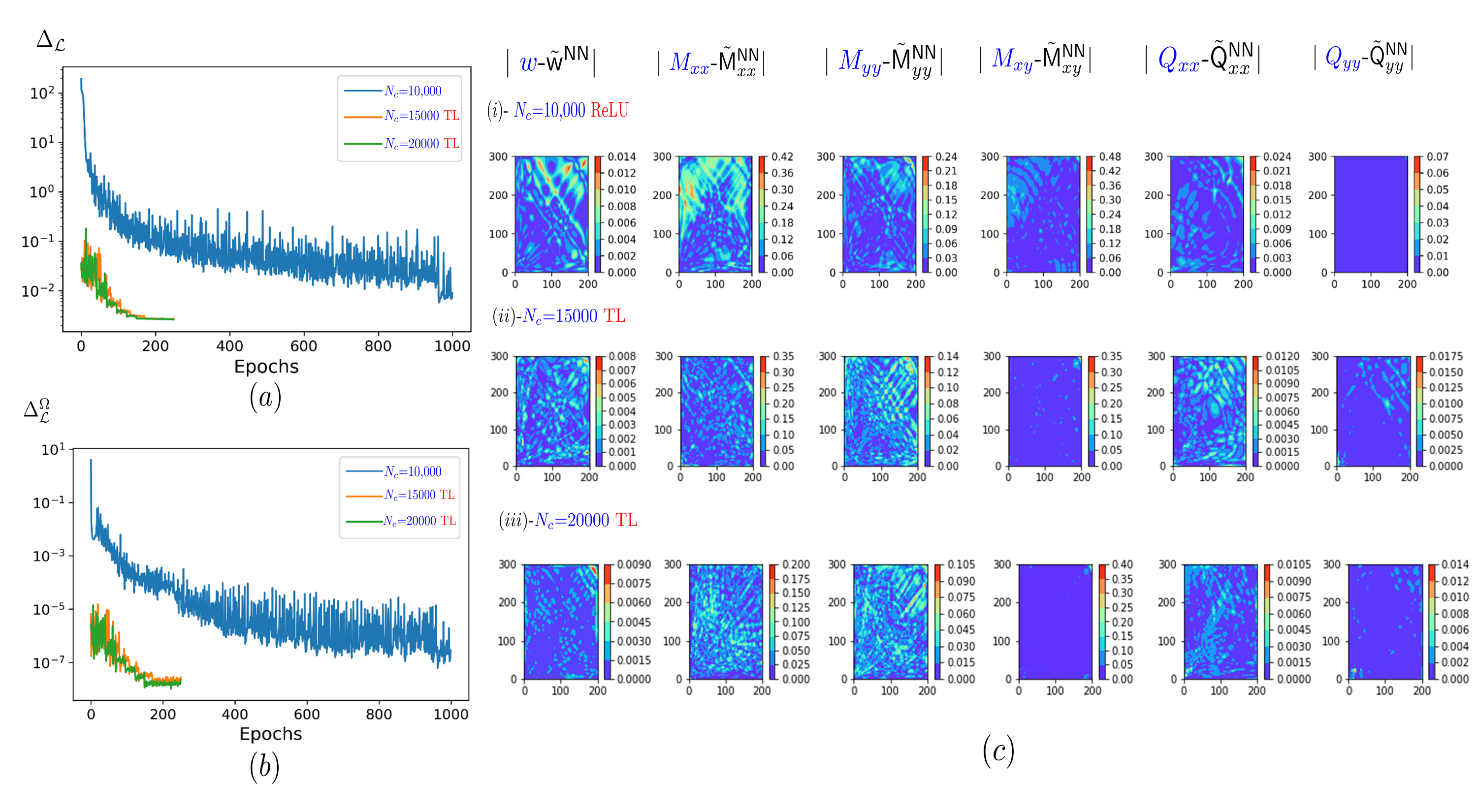}
\caption{\label{Fig-13} Influence of smart initialization of data-driven enhancement on (a) total loss $\Delta^{{\Omega}}_\mathcal{L}$; (b) constitutive loss  $\Delta^{{\Omega}}_\mathcal{L}$ for  increasing $N_c$ considering  ReLU activation; (c)  Absolute  error of field variables between analytical solution and  PINNs results for (i) $N_c=10,000$, (ii) $N_c=15,000$ TL, and $N_c=20,000$ TL.}
\end{figure}
\begin{table}
\centering
\caption{ Network parameters, training time, and the component of loss for different smart initialization of data-driven enhancement models.}
\begin{tabular}{c c c c c c c c c c c }
\\[-0.5em]
\hline
\\[-0.8em] 
\vtop{\hbox{\strut Network  }\hbox{\strut identifier}} & $N_c$  & Epochs  & $\Delta^{{\Omega}}_\mathcal{L}$  & $\Delta^{{\Gamma_t}}_\mathcal{L}$ & $\Delta^{{\Gamma_u}}_\mathcal{L}$ &$\Delta^{w}_\mathcal{L}$ & $\Delta^{{ M}}_\mathcal{L}$ & $\Delta^{{Q}}_\mathcal{L}$ & $\Delta_\mathcal{L}$ & \vtop{\hbox{\strut $t_{tr}$ }\hbox{\strut  ($min$)}}
\\[-0.0em]
\hline
\\[-0.5em]
N-1 & 10000 & 1000 & 5.34  & 132.31 & 1672.91  & 278.43  & 498.76 & 101.36 & 2689.11 & 23.1 
\\
\\[-0.5em]
N-TL1 & 15000 & 250 & 0.025 & 1.31 & 17.34  & 1.43  & 13.11 & 9.89 & 43.11 & 5.1 
\\
\\[-0.5em]
N-TL2 & 20000 & 250 & 0.005 & 0.71 & 2.96  & 2.01  & 2.56 & 0.87 & 9.11 & 7.2   
\\
\\[-0.5em]
\hline
\end{tabular}
\label{T-5}
\end{table}
\\
\\
{\bf 5.4.3  Smart initialization of data-driven enhancement:}
\\
\\
\\
In this section, we explore the applicability of data-driven enhancement in the proposed PINNs framework to improve the accuracy of the solution. 
Initially, the network is trained with relatively low $N_c=10,000$. 
The pre-trained model is then trained for the higher number of collocation datasets $N_c=15,000$ and $N_c=20,000$ to further improve the model accuracy. 
The idea is to speed up the training by utilizing pre-trained weights; the initial states of the PINNs models in the later phases of training are not random anymore. 
The speed-up is reflected in Figs. \ref{Fig-13}-(a, b) when the convergence of the loss curves ($\Delta_\mathcal{L}$ and $\Delta^{{\Omega}}_\mathcal{L}$) for the pre-trained models corresponding to $N_c=15,000$ and $N_c=20,000$ are much improved compared to the first training phase with $N_c=10,000$. 
In  Fig. \ref{Fig-13}-(c), the absolute errors between the approximations and analytical solutions are shown which demonstrate significant improvement of the PINNs approximations with the increase in $N_c$. 
Additionally, parameters related to the efficiency of the network training processes with initialization of data-driven enhancement are reported in Tab. \ref{T-5}. 
The loss terms quickly reduce by orders of magnitude in the second training phase which indicates that for the considered network architecture, $N_c=15000$ is possibly optimal. 
\\
\\
{\bf 6. Discussions  :}
\\
\\
In the current study, a generalized PINNs framework for solving problems in linear continuum elasticity in the field of solid mechanics is presented. 
The fundamentals of the PINNs framework involve a construction of the loss function for physics-informed learning of the NNs through the embedding of the linear constraint during training. 
Following the PINNs philosophy to solve the linear elastic problem accurately, a multi-objective loss function has been formulated and implemented. 
The proposed multi-objective loss function consists of the residual of the governing PDE, various boundary conditions, and data-driven physical knowledge fitting terms. 
Additionally, weights corresponding to the terms in the loss function dictate the emphasis on satisfying the specific loss terms. 
To demonstrate the efficacy of the framework, the Airy solution to an end-loaded cantilever beam and the Kirchhoff-Love plate theory governed by fourth-order Biharmonic PDE has been solved. 
The proposed PINNs framework is shown to accurately solve different fields in both problems. 
Parametric investigations on activation functions and network architectures highlight the scope of improvement in terms of solution accuracy and performance. 
Data-driven enhancement of the PINNs approximations using analytical solutions significantly boosts accuracy and speed only using minimal network parameters. 
Therefore, such an approach can be employed to enhance solution accuracy for complex PDEs. 
Additionally, the applicability of a smart initialization of data-driven enhancement learning-based approach quickening the training process and also improving model accuracy have been illustrated. 
Such an approach would be key in achieving computational efficiency beyond conventional computational methods for solving linear continuum elasticity.
The proposed PINNs elasticity solvers utilize Tensorflow as the backend which can be easily deployed in CPU/ GPU clusters, whereas, conventional algorithms lack such adaptability.  
Thus, it opens new possibilities for solving complex elasticity problems that have remained unsolved by conventional numerical algorithms in the regime of continuum mechanics.
It is however worth noting that exploitation of the computational advantages of the PINNs framework depends on various factors including the choice of the network architectures, hyperparameter tuning, sampling techniques (distribution) of collocation points, etc. 
It has been shown that appropriate combinations of such factors significantly improve the training process and the trained models.

In the present study, random sampling of the collocation points has been considered which is simple, yet powerful, that can lead to a significantly better reconstruction of the elastic fields. Importantly, this approach does not increase computational complexity, and it is easy to implement. However, in elastic/elastoplastic PDE problem which exhibits local behavior (e.g., in presence of sharp, or very localized, features) or problems with singularities the performance of PINNs may vary drastically with various sampling procedures \citep{daw2022rethinking,leiteritz2021avoid}. To overcome such an issue, a failure-informed adaptive enrichment strategy such as failure-informed PINNs (FI-PINNs) can be employed that adopts the failure probability as the posterior error indicator to generate new training points in the failure region \citep{gao2022failure}. Furthermore, the basic resampling scheme can be further improved with a gradient-based adaptive scheme to relocate the collocation points through a cosine-annealing to areas with higher loss gradient, without increasing the total number of points that demonstrated significant improvement under relatively fewer number of collocation points and sharper forcing function \citep{subramanian2022adaptive}. In addition, the evolutionary sampling (Evo) method \citep{daw2022rethinking}  that can incrementally accumulate collocation points in regions of high PDE residuals can be an efficient choice for solving various time-dependent PDEs with little to no computational overhead. Instead of using a random approach such as Latin Hypercube sampling, in the future, different deterministic and pseudo-random sampling strategies such as Sparse Grid sampling or Sobol Sequences can be employed to further improve the performance of the model.

Furthermore, it is critical to obtain the statics of saturation along different parts of the solution domain during the training of DNNs \citep{glorot2010understanding,rakitianskaia2015saturation}. The saturation occurs when the hidden units of a DNN predominantly output values close to the asymptotic ends of the activation function range which reduces the particular PINNs model to a binary state, thus limiting the overall information capacity of the NN \citep{rakitianskaia2015measuring,bai2019adaptive}. The saturated units can make gradient descent learning slow and inefficient due to small derivative values near the asymptotes which can hinder the training PINNs efficiently \citep{bai2019adaptive}. Thus, in the future, NN saturation can be studied quantitatively in relation to the ability of NNs to learn, generalize, and the degree of regression accuracy. 
In addition, various weighting coefficients of the loss terms in Eq. \ref{E-8} and implementation of second-order optimization techniques \citep{tan2019review} can accelerate the training significantly. 
Based on the performance of the PINNs framework herein, further studies quantifying the computational gains of the PINNs approach compared to conventional numerical methods are in order.
The proposed approach can be extended to the solution in various computational mechanics problems such as soil plasticity \citep{chen1985soil,bousshine2001softening}, strain-gradient plasticity \citep{guha2013finite,guha2014fracture}, composite modeling \citep{roy2021finite} etc. 
Furthermore, the present model can be employed to predict microstructure evolution in Phase-field (PF)  approach  including  various solid-solid phase transitions (PTs) 
\citep{levitas2013multiple,levitas2015multiphase,roy2020influence,roy2020effects,roy2020evolution}, solid-solid PT via intermediate melting \citep{levitas2016multiphase,roy2021barrierless,roy2021multiphase,roy2021influence,roy2021formation,roy2021energetics,roy2022multiphase}, etc.
\\
\\
{\bf 7. Conclusions  :}
\\
\\
Summarizing, the current work presents a deep learning framework based on the fundamentals of  PINNs theory for the solution of linear elasticity problems in continuum mechanics. 
A multi-objective loss function is proposed for the linear elastic solid problems that include governing PDE, Dirichlet, and Neumann boundary conditions across randomly chosen collocation points in the problem domain. 
Multiple deep network models trained to predict different fields result in a more accurate representation. 
Traditional ML/ DL approaches that only rely on fitting a model that establishes complex, high-dimensional, non-linear relationships between the input features and outputs, are unable to incorporate rich information available through governing equations/ physics-based mathematical modeling of physical phenomena. 
Conventional computational techniques on the other hand rely completely on such physical information for prediction. 
The PINNs approach combines the benefits of the DL techniques in the extraction of complex relations from data with the advantages of the conventional numerical techniques for physical modeling. 
The proposed method may be extended to nonlinear elasticity, viscoplasticity,  elastoplasticity, and various other mechanics and material science problems. 
The present work builds a solid foundation for new promising avenues for future work in machine learning applications in solid mechanics. 
\\
\\
{\bf Acknowledgements:}
The support of the Aeronautical Research and Development Board (Grant No.
DARO/08/1051450/M/I) is gratefully acknowledged.
\\
\\
{\bf Competing interests:}
The author declares no competing interests.


\bibliographystyle{apalike}




\bibliography{Reference_PINNS} 


\end{document}